\documentclass[runningheads]{llncs}
\usepackage{perpage} %the perpage package
\MakePerPage{footnote}
\usepackage{graphicx}
\usepackage{subcaption}
\usepackage{tikz}
\usepackage{multirow}
\usepackage{comment}
\usepackage{amsmath,amssymb} % define this before the line numbering.
\usepackage[mathscr]{euscript}
\usepackage{color}
\usepackage[accsupp]{axessibility}  % Improves PDF readability for those with disabilities.

\usepackage[width=122mm,left=12mm,paperwidth=146mm,height=193mm,top=12mm,paperheight=217mm]{geometry}

\begin{document}
% \linenumbers
\pagestyle{headings}
\mainmatter
\def\ECCVSubNumber{2100}  % Insert your submission number here

\title{Scalable Vehicle Re-Identification via Self-Supervision} % Replace with your title

\begin{comment}
\titlerunning{} 
\authorrunning{} 
\author{}
\institute{}
\end{comment}
%******************

%\begin{comment}
\titlerunning{Scalable Vehicle Re-Identification via Self-Supervision}
\author{Pirazh Khorramshahi \and
Vineet Shenoy \and
Rama Chellappa}
\authorrunning{P. Khorramshahi et al.}
\institute{Artificial Intelligence for Engineering and Medicine Lab, \\ Johns Hopkins University, Baltimore, MD \\
\url{https://aiem.jhu.edu} \\
\email{\tt\small \{pkhorra1, vshenoy4, rchella4\}@jhu.edu}}
%\end{comment}
%******************
\maketitle

\begin{abstract}
As Computer Vision technologies become more mature for intelligent transportation applications, it is time to ask how efficient and scalable they are for large-scale and real-time deployment. Among these technologies is Vehicle Re-Identification which is one of the key elements in city-scale vehicle analytics systems. Many state-of-the-art solutions for vehicle re-id mostly focus on improving the accuracy on existing re-id benchmarks and often ignore computational complexity. To balance the demands of accuracy and computational efficiency, in this work we propose a simple yet effective hybrid solution empowered by self-supervised training which only uses a single network during inference time and is free of intricate and computation-demanding add-on modules often seen in state-of-the-art approaches. Through extensive experiments, we show our approach, termed Self-Supervised and Boosted VEhicle Re-Identification (SSBVER), is on par with state-of-the-art alternatives in terms of accuracy without introducing any additional overhead during deployment. Additionally we show that our approach, generalizes to different backbone architectures which facilitates various resource constraints and consistently results in a significant accuracy boost.

\end{abstract}
\section{Introduction}
The problem of vehicle re-identification (re-id) is essentially a retrieval task in which a query vehicle image is presented and correct image matches to the query identity are retrieved from a large gallery set. The gallery set is composed of large number of vehicle images that are captured at different times of day, from traffic cameras mounted at different locations and under varying weather conditions. Therefore, the vehicle re-id task becomes quite challenging as a given vehicle's appearance can drastically vary under different viewpoints, camera and lighting conditions. On the other hand, many vehicles can appear similar due to relatively small variations in vehicle manufacturers, models, trims, years and colors. To address this task and its associated challenges, discriminative visual representation learning via Deep Neural Networks (DNNs) has become the de facto approach. Note that vehicle re-id is objectively different than vehicle classification task where the goal is to identify a vehicle's model rather than its instance. Therefore, vehicle re-id requires more fine-grained features, particularly within local regions, to highlight the differences in similar looking vehicles. As a result, a significant number of research works have been undertaken to develop attention mechanisms into the DNNs' pipeline in both implicit \cite{hu2018squeeze,vaswani2017attention,woo2018cbam} and explicit ways \cite{He_2019_CVPR,khorramshahi2019dual}. While these approaches are successful in improving the state-of-the-art, they often require rich data annotations and demand heavy computation that raise scalability issues. The burden of deploying such models in real-time applications such as city-scale multi-camera tracking quickly becomes evident as hundreds of traffic cameras \footnote{\url{https://trafficview.org}} should be processed simultaneously under limited computational resources. In addition, each camera can potentially capture tens of vehicles each second for which visual representations should be computed. We also note that the dimensionality of representations computed by re-id models is another important aspect that should be taken into account. Within a multi-camera tracking system, these representations are transported across different processes for both single- and multi-camera tracking purposes and can become unmanageable if dimensionality is too large. Consequently, it is paramount that a vehicle re-id module benefits from an efficient design that can effectively learn discriminative representations from vehicle re-id datasets without relying on the existence of additional annotations beyond ID labels, \emph{e.g.} vehicle's manufacturer, model, color, key-points or parts' location. This poses the following question: \textit{How can we learn more robust representation of vehicles' images using efficient DNN-based models and without the incorporation of additional labels?} 

Recently, there have been great strides in the area of Self-Supervised Learning (SSL) particularly for the task of image classification to learn robust embeddings without the incorporation of human-generated labels. As a result, the performance gap between self-supervised and fully-supervised learning has become narrower. In addition, SSL methods outperform mainstream supervised pretraining when transferred to down-stream tasks such as object detection and demonstrate better data efficiency \cite{chen2020simple,he2020momentum,caron2020unsupervised,grill2020bootstrap}. This has motivated us to explore the viability of recent self-supervised learning techniques in the context of vehicle re-id. 
A great number of recent works in SSL classify \cite{NIPS2014_07563a3f} or discriminate \cite{he2020momentum,chen2020simple} each image as a separate class known as \textit{Instance Classification} and \textit{Instance Discrimination} respectively via contrastive learning. While these approaches yield robust representations for the image classification task, there is no clear path to extend them to object re-identification where there are multiple images corresponding to the same ID which should not be discriminated against one another. To address this issue, supervised contrastive learning \cite{khosla2020supervised}, a generalization of Triplet loss \cite{weinberger2009distance}, has been proposed so that similar images are considered as positives during training. This is identical to the current practice in object re-id which employs triplet loss as standard. In contrast, the recent SSL method, namely DINO \cite{caron2021emerging} casts the SSL as self-distillation and establishes the connection between Knowledge Distillation and SSL in the absence of labels without performing any discriminatory task among images. As we see in section \ref{sec:method}, this creates the opportunity to enrich the learning of a re-id model with the self-supervisory signal and encourages the local to global correspondence, essentially  mimicking the attention mechanism. As discussed in section \ref{sec:experiments}, DINO alleviates the intra-class variation through feature compactness. 

The contributions of our work can be summarized as the following:
\begin{itemize}
    \item [1-] Introduction of self-supervised representation learning directly to the training pipeline of vehicle re-id.
    \item [2-] Presenting a simple, efficient and highly accurate baseline without carrying any additional overhead in the inference phase.
    \item [3-] Evaluation of the proposed approach on recent DNN architectures including SWIN Transformer \cite{liu2021Swin} and Convnext \cite{liu2022convnet}.
\end{itemize}

The rest of the paper is organized as follows. In section \ref{sec:related_work}, we review recent works in the area of vehicle re-id. The proposed method and its detailed architecture is discussed in section \ref{sec:method}. Through extensive experimentations in section \ref{sec:experiments}, we show the effectiveness of our approach on multiple challenging vehicle re-id benchmarks and with different backbone architectures, obtaining state-of-the-art results in terms of accuracy-efficiency trade-off. Finally, in section \ref{sec:ablation} we further analyze SSBVER and validate our design choices. Section \ref{sec:conclusion} concludes the paper.

\section{Related Work}
\label{sec:related_work}
In recent years, vehicle re-identification has attracted a significant amount of attention thanks to its critical role in the development of smart transportation technologies. Here we review a number of selected works that has been published in recent years. 

Learning discriminative features for vehicles demands curated datasets of vehicles' images of diverse makes, models, colors with high number of identities. To this end, several datasets have been introduced over the past several years which contributed to the current landscape of vehicle re-id. Among these are VeRi \cite{liu2016large}, VehicleID \cite{liu2016deep}, VeRiWild \cite{lou2019large}, Vehicle1M \cite{guo2018learning}, PKU VD \cite{yan2017exploiting}, and CityFlow Re-id \cite{Tang19CityFlow}. Each of these datasets has different attributes and variations in terms of scale, resolution; however, only VeRi, VeRiWild, and CityFlow Re-id capture vehicles from diverse views that is more representative of unconstrained vehicle re-id. Additionally, a video-based vehicle re-id dataset, namely VVeRI-901 is curated by \cite{zhao2021phd} to motivate research towards the incorporation of temporal information for representation learning. Since vehicle re-identification is concerned with subtle cues and small-scale details on vehicle images as opposed to vehicle classification task, authors in \cite{wang2017orientation} annotated images in the VeRi dataset with view point labels and key-point information such as the location of logo, head and tail lights, side mirrors and corners. This provides opportunities to devise supervised attention models to adaptively extract local features based on vehicle's orientation and obtain a more discriminative embedding \cite{khorramshahi2019dual,khorramshahi2019attention}. Similarly, \cite{He_2019_CVPR} annotated VehicleID dataset images with parts' bounding box information to detect and extract fine-grained features of vehicle images. While having extra annotations help to learn where to look for discriminative information, it is not a scalable approach. To address this issue, authors in \cite{khorramshahi2020devil,Peri_2020_CVPR_Workshops} developed a variational auto-encoder model to reconstruct vehicle images in a coarse manner and obtain self-supervised saliency maps highlighting identity-dependant information to either adjust vehicle images directly or excite intermediate features maps of the underlying DNN. Similarly, authors in \cite{li2021self} proposed a self-supervised model based on the pretext task of image rotation to learn geometric features along with appearance information. As orientation is one of the factors that can negatively bias the learned embeddings of a re-id system, authors of \cite{meng2020parsing} and \cite{bai2020disentangled} propose to learn view-aware aligned features and to disentangle the orientation from visual features respectively. To extract region-specific features, a heterogeneous relational graph-based model has been introduced in \cite{zhao2021heterogeneous} to encode the relation of the different local regions into a unified representation. These methods are mainly designed to improve the re-id accuracy without any consideration for practical issues and efficiency metrics as discussed in section \ref{sec:sota}. Authors in \cite{mcmt} discuss the importance of speed and memory efficiency for real-time and large-scale multi-camera tracking as a transportation application. Therefore, we present SSBVER, a hybrid approach that employs the power of self-supervision to boost the performance of vehicle re-id without any additional overhead during inference time. 

\section{Method}
\label{sec:method}
\begin{figure}[ht]
    \centering
    \includegraphics[width=0.6\textwidth]{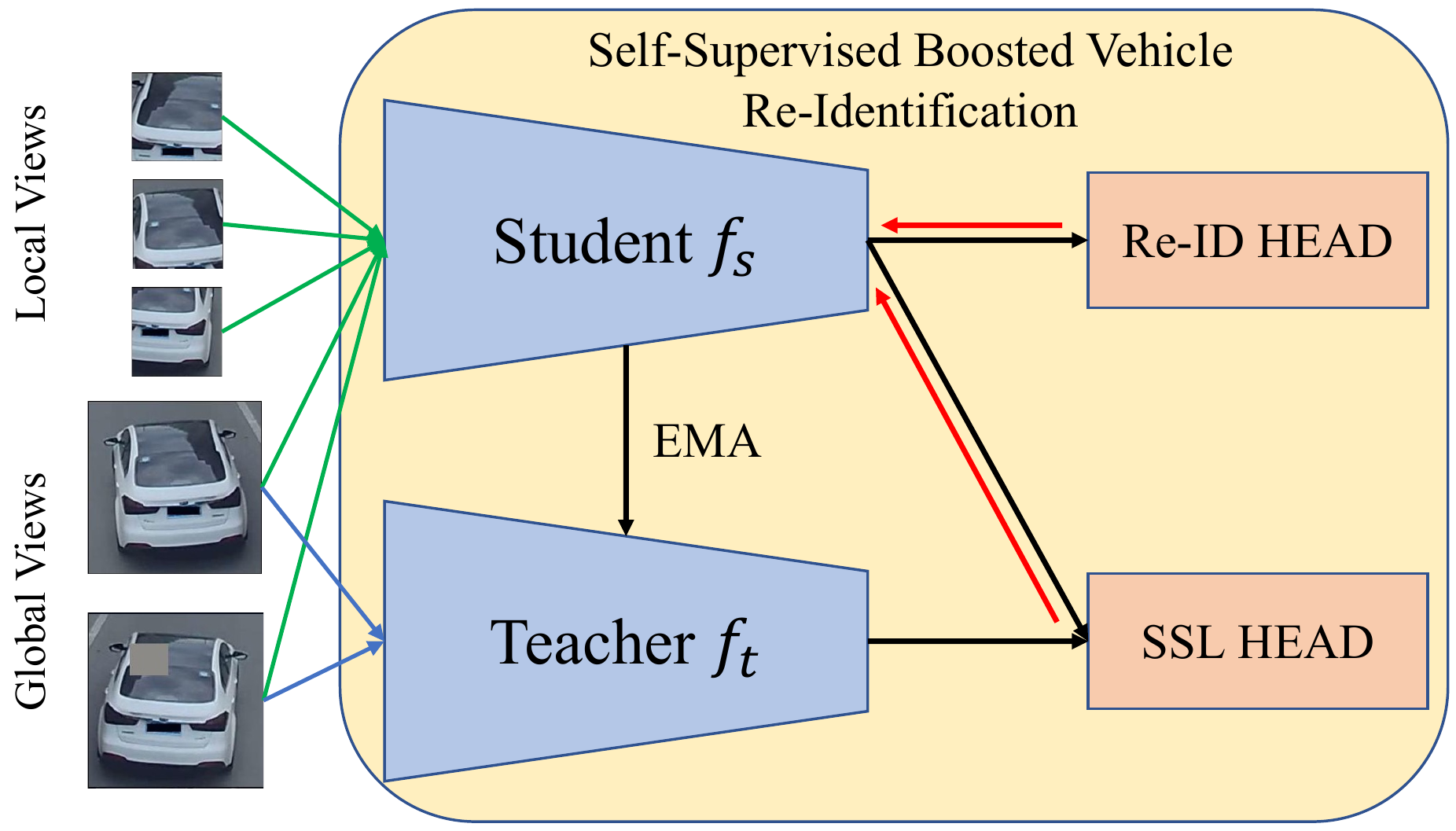}
    \caption{Self-supervised Boosted Vehicle Re-identification Pipeline. Only the student model is optimized for re-identification (Re-ID Head) and self-supervision (SSL Head) objectives. The teacher model is obtained by taking the exponential moving average (EMA) of student model over the course of training.}
    \label{fig:pipeline}
\end{figure}

In this section we discuss the details of the proposed Self-Supervised Boosted Vehicle Re-identification pipeline shown in Fig. \ref{fig:pipeline}.
\subsection{Backbone Feature Extractors}
Inspired by recent SSL methods, our approach benefits from a student and teacher pairing, where both the student and the teacher have identical architectures. The choice of the architecture is arbitrary and can be selected based on the application and resource constraints. In this work, we adopt multiple candidate architectures including ResNet \cite{he2016deep}, ResNet\_IBN \cite{pan2018IBNNet}, Vision Transformer (ViT) \cite{dosovitskiy2020image}, SWIN Transformer \cite{liu2021Swin}, and ConvNext \cite{liu2022convnet} to study the generalization capability of SSBVER. The teacher model is considered a momentum encoder as it is a low-pass version of the student model via taking the exponential moving average over the course of training iterations with the momentum parameter $\lambda$, \emph{i.e.} ${\theta_t}^i = \lambda{\theta_t}^{i-1} + (1-\lambda){\theta_s}^{i}$ where $\theta_t$, $\theta_s$ and $i \ge 1$ are teacher model parameters, student model parameters, and the current training iteration respectively. Note that both models are initialized from the same set of ImageNet pre-trained weights, \emph{i.e.} $\theta_t^0 = \theta_s^0$.
\subsection{Re-Identification Head}
\begin{figure}
    \centering
    \includegraphics[width=0.6\textwidth]{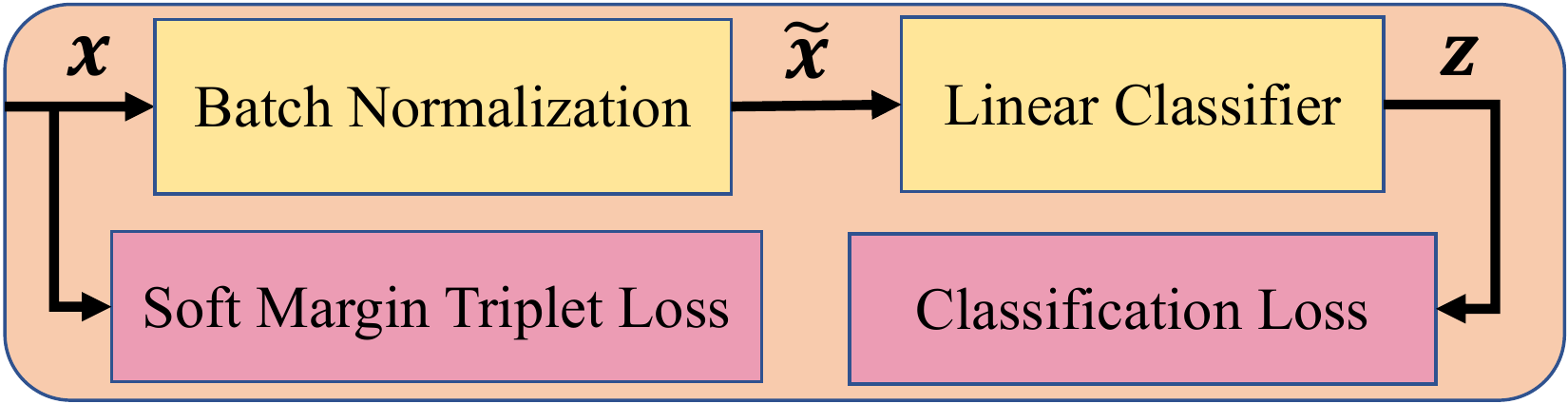}
    \caption{Re-Id Head: Extracted visual features $x$ from the student model $f_s$ are passed through a bottleneck layer implemented by 1-dimensional Batch Normalization to obtain feature $\Tilde{\mathbf{x}}$. Subsequently, classification logits $\textbf{z}$ are obtained from a linear classifier. Soft-margin Triplet and Cross Entropy loss functions constrain $\mathbf{x}$ and $\mathbf{z}$ respectively.}
    \label{fig:reid_head}
\end{figure}

SSBVER uses the re-identification head to constraint the extracted features $\mathbf{x}$ by the backbone student model so that those corresponding to the same identity are embedded close together while the ones belonging to different identities kept  apart. This goal is realized by employing Triplet loss function in conjunction with Cross Entropy loss function and results in a strong baseline model as demonstrated in prior works \cite{luo2019bag,he2020fastreid,khorramshahi2020devil,Peri_2020_CVPR_Workshops,khorramshahi2021towards}. Fig. \ref{fig:reid_head} outlines the inner workings of the re-identification head. The soft-margin triplet loss with batch-hard sampling method is computed via the following formulation:
\begin{equation}
    \label{eq:soft_triplet}
    \centering
    \small{
    \mathcal{L}_{t} = \sum_{a \in b_{i}} \log \left(1 + \exp(\max_{p \in \mathcal{P}(a)} {||\mathbf{x}_a-\mathbf{x}_p||}_2 - \min_{n \in \mathcal{N}(a)} {||\mathbf{x}_a-\textbf{x}_n||}_2 )\right)
    }
\end{equation}
In Eq. \ref{eq:soft_triplet},  $b_i$ denotes the $i^{th}$ training batch. In addition, $a$, $\mathcal{P}(a)$ and $\mathcal{N}(a)$ are an anchor sample and its corresponding positive and negative sets defined within batch $b_i$ accordingly. Note that in the formulation of soft-triplet loss function, there is no notion of a pre-defined margin which is heuristically set in prior works and adds to the number of hyper-parameters. Once the representation vector $x \in \mathbb{R}^d$ is computed, it is passed to a batch normalization layer to obtain $\Tilde{\mathbf{x}}$ with $\Tilde{\mathbf{x}}_i = \frac{\mathbf{x}_i - E[\mathbf{x}_i]}{\sqrt{Var(\mathbf{x}_i)}}$. $E(\mathbf{x}_i)$ and $Var(\mathbf{x}_i)$ are the mean and variance of $i^{th}$ dimension that are computed across a batch. Authors in \cite{luo2019bag} showed that employing this bottleneck layer helps the consistency of Triplet and Cross Entropy classification loss functions in the context of re-id. Afterwards, the linear classifier computes the class logit vector $\mathbf{z} \in \mathbb{R}^{k}$ ($k$ is the total number of training IDs) through the linear operation $\mathbf{z} = W\Tilde{\mathbf{x}} + B$. $W \in \mathbb{R}^{k \times d}$ and $B \in \mathbb{R}^{k}$ are the weight matrix and bias of the classifier correspondingly. The classification loss is computed as follows: 
\begin{equation}
    \label{eq:cross_entropy_cls}
    \centering
    \mathcal{L}_{c} = -\sum_{j=1}^{k} y^i_j\log\hat{y}^i_j, \quad \hat{y}^i_j = \frac{e^{z_j^{i}}}{\left(\sum_{m=1}^{k}e^{z_m^{i}}\right)}
\end{equation}
In Eq. \ref{eq:cross_entropy_cls}, $\hat{y}_j^i$ is the probability that the $i^{th}$ sample belongs to the class $j$. In addition, we employ label-smoothing as a regularization method following the work of \cite{szegedy2016rethinking}. Therefore, rather than considering the ground-truth vector as a one-hot encoding vector, it is computed as:
\begin{equation}
    \label{eq:gt_vector}
    \centering
    y_j^i = \begin{cases} 
    1 - \frac{k-1}{k}\epsilon & j = k(i) \\ 
    \frac{\epsilon}{k} & \text{otherwise}
    \end{cases}
\end{equation}
where $\epsilon \in [0,1]$ and $k(i)$ are a hyper-parameter and the class label of the $i^{th}$ sample.

While optimizing Triplet and Cross Entropy classification loss functions on the extracted representations results in a strong and efficient baseline re-id model, the incorporation of attention mechanisms to focus on local regions of vehicle images and extract subtle cues can further improve the performance. However this improvement is achieved at the expense of increased complexity and computation time that can be prohibitive when applied to large-scale and real-time scenarios. To overcome this shortcoming with the goal of minimizing inference complexity while enjoying enhanced accuracy, we incorporate SSL in the training phase of the re-id model to encourage the local to global correspondence, mimicking the attention mechanism. This is discussed in the following section.

\subsection{Self-Supervised Learning Head}

\begin{figure}
    \centering
    \includegraphics[width=0.6\textwidth]{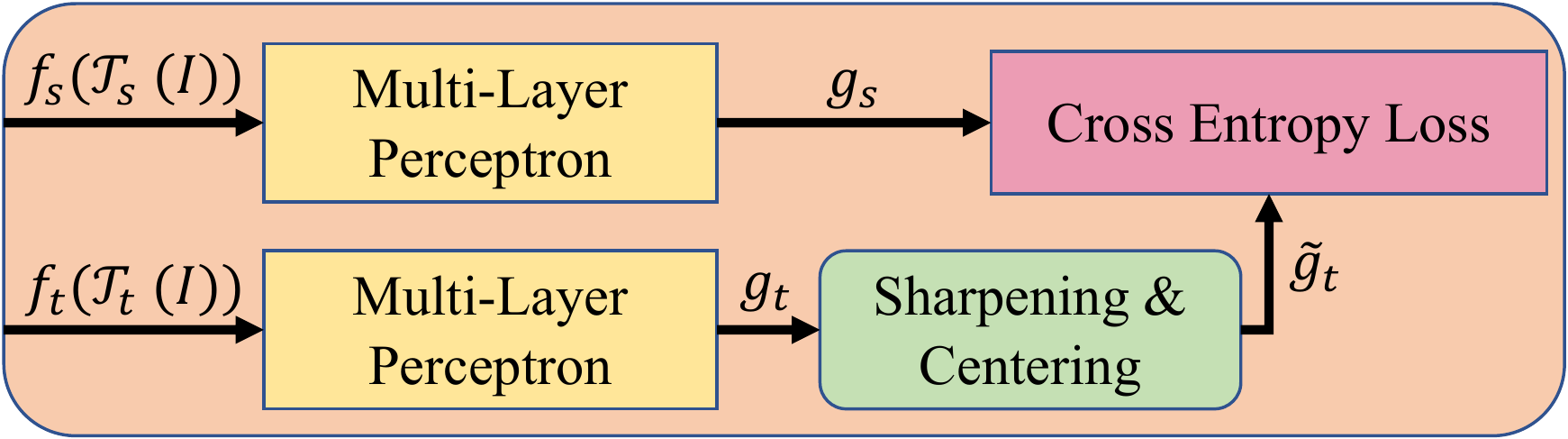}
    \caption{Self-Supervised Learning Head: Inspired by self-distillation with no labels (DINO) \cite{caron2021emerging}, a vehicle's image $I$ is randomly augmented using $\mathscr{T}_s$ and $\mathscr{T}_t$ transformations and is separately passed to student and teacher backbone models followed by multi-layer perceptron modules to obtain $g_s$ and $g_t$ prediction vectors respectively. After applying sharpening and centering operations to the teacher's model output to avoid collapse, the cross entropy of student and teacher predictions is minimized.}
    \label{fig:dino_head}
\end{figure}

To improve the performance of re-id without the incorporation of any additional annotation on vehicles' parts and attributes, we propose to apply a self-supervised optimization objective based on self-training and knowledge distillation over the course of training. Self-distillation with no labels (DINO) \cite{caron2021emerging} presents a self-supervised learning paradigm with multi-crop strategy \cite{he2020momentum} in which semantically rich general representations can be learned from scratch and demonstrates competitive performance when transferred to down-stream tasks. In contrast to most recent SSL approaches, DINO does not solve the instance discrimination task and therefore does not rely on any negative sampling. This makes DINO a fitting choice for re-id as each identity can be represented by multiple images that should not be discriminated against each other. 

After sampling a vehicle's image $I$, we create two sets of views on the fly, namely $V_g(I)$ and $V_l(I)$, where $V_g(I) = \{I_{g_{1}},I_{g_{2}}\}$ contains two different global views and $V_l(I) = \{I_{l_{1}},\dots,I_{l_{L}}\}$ has $L$ local views of image $I$. Images in $V_g(I)$ are generated by $\mathscr{T}_t$ with randomly cropping a region of image with random area ratio $a_g$, padding zeros on the edges, flipping horizontally, jittering colors and erasing a random patch to simulate occlusion \cite{hermans2017defense}. To generate images in $V_l(I)$, $\mathscr{T}_s$ crops a random portion of image $I$ with random area ratio of $a_l$, randomly flips and jitters color. The teacher only processes images in $V_g(I)$ while the student model is fed by images in both sets \emph{i.e.} $V_g(I) \cup V_l(I)$. As shown in Fig. \ref{fig:dino_head}, after obtaining representations $\mathbf{x}_s = f_{s}(\mathscr{T}_s(I))$ and $\mathbf{x}_t = f_{t}(\mathscr{T}_t(I))$, they are mapped to another space using multi-layer perceptrons (MLP) with four hidden layers, and Gaussian Error Linear Units (GELU) \cite{hendrycks2016gaussian} to yield $E$-dimensional vectors $g_s$ and $g_t$. A common problem that is associated with SSL-based approaches with a pair of networks, is the issue known as collapse where both encoders learn to output trivial embeddings irrespective of the input images to minimize the loss function. There has been a number of techniques to prevent collapse including contrastive learning with negative pairs \cite{chen2020simple}, stop-gradient \cite{chen2021exploring}, clustering \cite{caron2020unsupervised}, momentum encoder \cite{grill2020bootstrap}, and redundancy reduction of the learnt representation's dimensions \cite{zbontar2021barlow}. DINO is optimized by minimizing the cross entropy loss between student and teacher models' output so that the student model can match the teacher's prediction. While it uses momentum encoder and stop gradient techniques to battle collapse, it is shown collapse can still occur in the form of either outputting uniform predictions or having a single dimension to dominate others regardless of the inputs. To counteract, centering and sharpening of the teacher's outputs are proposed \cite{caron2021emerging}. In centering, an exponential moving average $\mathbf{c}$ of teacher's predictions is recorded and subtracted from its predictions to prevent the domination of a single dimension. On the other hand, a relatively small temperature is applied to the teacher's results in the softmax function to battle the uniformity of the outputs. Therefore sharpening and centering operations attempt to establish a balance in which collapse does not occur. This cross entropy loss is calculated with the following formulation:
\begin{equation}
    \label{eq:dino}
    \centering
    \mathcal{L}_s = - \sum_{\scalebox{0.6}{$I\in V_g(I)$}} \sum_{\begin{array}{c}
                            \scalebox{0.6}{$I^{'}\in V_g(I) \cup V_l(I)$} \\
                            \scalebox{0.6}{$I^{'} \neq I$} \end{array}} \sum_{i=1}^{E}
                            p_t^i(I)\log(p_s^i(I^{'}))
\end{equation}
Where $p_s^i(I^{'}) = \frac{\exp(g^i_s(f_s(I^{'}))/\tau_s)}{\sum_{j}\exp(g^j_s(f_s(I^{'}))/\tau_s)}$ and $p_t^i(I) = \frac{\exp((g^i_t(f_t(I)) -c^i)/\tau_t)}{\sum_{j}\exp((g^j_t(f_t(I)) - c^j)/\tau_t)}$. Also $\tau_s$, $\tau_t$ are student and teacher models' softmax temperatures. $c^i$ is the $i^{th}$ element of the vector $\mathbf{c}$ that is the exponential moving average of teacher's predictions $g_t$. Note that the cross entropy is only calculated when student and teacher are processing different augmented versions of an image, \emph{i.e.} $I^{'} \neq I$. 

\subsection{End-to-End Training}
In our experiments, we first establish a baseline model setup in which only $\mathcal{L}_c$ and $\mathcal{L}_t$ are used to train the student model. The teacher model which is the exponential moving average of the student model over the training iterations is used for evaluation. Afterwards, the setup of the SSBVER outlined in Fig. \ref{fig:pipeline} is used for model training and the total loss function for  end-to-end training is calculated as follows:
\begin{equation}
    \label{eq:total_loss}
    \centering
    \mathcal{L}_{total} = \lambda_c \mathcal{L}_c + \lambda_t \mathcal{L}_t + \lambda_s \mathcal{L}_s
\end{equation}
In Eq. \ref{eq:total_loss}, the coefficients $\lambda_c$, $\lambda_t$, and $\lambda_s$ are the weights corresponding to each of the loss terms and are empirically set. We emphasize that the gradients of the loss functions in Eqs. \ref{eq:dino}, \ref{eq:cross_entropy_cls}, \ref{eq:soft_triplet} are computed with respect to only student model's parameters $\theta_s$.
\section{Experimental Results}
\label{sec:experiments}
To evaluate the proposed SSBVER method and understand how much it can benefit the re-identification task without introducing any additional overhead during test time, we use the three widely used VeRi \cite{liu2016large}, VehicleID \cite{liu2016deep} and VeRiWild \cite{lou2019large} datasets. Additionally, we use ResNet \cite{he2016deep}, ResNet\_IBN \cite{pan2018IBNNet}, Vision Transformer (ViT) \cite{dosovitskiy2020image}, SWIN Transformer \cite{liu2021Swin} and Convnext \cite{liu2022convnet} backbone feature extractor models to study the extent to which SSBVER generalizes to different model architectures. In this section, we discuss the three vehicle re-id datasets, evaluation metrics, implementation details, and finally present our experimental results.

\subsection{Datasets}
\textbf{VeRi} \cite{liu2016large} is the first multi-view real-world vehicle re-id dataset. It is regarded as a large-scale dataset; however, compared to the size of more recent datasets it is relatively small. The training and testing sets contain $37,778$ and $13,257$ images of $576$ and $200$ vehicle identities respectively.  \\

\noindent
\textbf{VehicleID} \cite{liu2016deep} is a comparatively larger benchmark as it contains $113,346$ ($108,417$) images of $13,164$ ($13,103$) unique vehicles in the training (testing) set. In contrast to VeRi, images in Vehicle ID are mainly captured from either front or rear of vehicles which impacts the dataset's representativeness of the real-world scenarios. For evaluation, multiple splits of different sizes are created from the original test set and referred to as small medium and large which contain $800$, $1600$, and $2400$ unique identities.\\

\noindent
\textbf{VeRiWild} \cite{lou2019large} with $416,314$ images of $40,671$ individual identities is the largest multi-view vehicle re-id dataset in the wild that is captured via $174$ traffic cameras and have variations in lighting and weather conditions. The training set contains $277,797$ images of $30,671$ identities and test set, similar to VehicleID dataset, is split into three small, medium and large sets of $3000$, $5000$, and $10,000$ unique identities.

\subsection{Evaluation Metrics}
\label{subsec:reid_metric}
The following evaluation metrics is widely adopted in the re-id community to measure the success of re-id systems. To this end, upon receiving a query image, visual representations are computed for query and the entire gallery. Afterwards, a distance measure, \emph{e.g.} Euclidean or Cosine, is used to computed the similarity scores and rank the gallery.

\noindent
\textbf{Mean Average Precision (mAP)} shows how well the gallery is ranked with respect to the query image. All the corresponding true matches to the query identity participate in the calculation of mAP.

\noindent
\textbf{Cumulative Match Curve (CMC) @ $k$} yields the probability that there exists at least one correct match to the query image in the top $k$ items in the ranked gallery. Note that this metric for $k=1$, \emph{i.e.} CMC@$1$ is valuable for designing a multi-camera tracking system since association of tracks across cameras is highly dependant on the pairs with highest similarity score. 

\subsection{Implementation Details}
As mentioned before we first setup a baseline model upon which we build the SSBVER idea and compare the experimental results accordingly. To ensure that SSBVER and baseline experiments are comparable, in the baseline experiments we record the exponential moving average with momentum parameter $\lambda=0.9995$ of the feature extractor's parameter, \emph{i.e.} to mimic the teacher (momentum encoder) in SSBVER, which is used for evaluation. The total training epochs is set to $120$ for all models and datasets. Label smoothing parameter is set to $\epsilon=0.2$ in the classification objective $\mathcal{L}_c$. To create global and local views every time $a_g \in [0.8,1]$ and $a_l \in [0.1, 0.4]$ are randomly selected. For ResNet and ResNet\_IBN architectures, we use the Adam \cite{kingma2014adam} optimizer, learning rate of $\eta=0.0005$ with Gamma decay factor $\gamma=0.1$ at $40^{th}, 70^{th}, 100^{th}$ epochs and weight decay of $0.001$. For ViT, SWIN and Convnext we use the \textit{base} model variant, AdamW \cite{loshchilov2017decoupled} optimizer and cosine learning rate decay scheduling \cite{loshchilov2016sgdr} with $\eta_{max}=0.0001$ and $\eta_{min}=1.6e-5$. In addition, weight decay is set to $0.0001$. Finally, linear learning rate warm-up \cite{fan2019spherereid} (with rate $0.099$) is adopted for the first ten epochs for all the experiments and model architectures. In the SSL head, the student's temperature is fixed and set to $\tau_s=0.1$ while the teacher's temperature $\tau_t$ is increased linearly from $0.0005$ to $0.001$ in the first ten epochs and remains fixed for the rest of training epochs. Additionally the output dimensionality of the SSL head $E$ is set to $1024$, $8192$ and $16384$ for the VeRi, VehicleID and VeRiWild datasets respectively. Using these hyper-parameters, we train baseline and SSBVER models for every choice of backbone architecture and dataset. In the next section, we present the evaluation metrics for these experiments and discuss how the performance of vehicle re-id improves by adopting self-supervision and how the choice of backbone architecture further impacts the overall performance. 

\subsection{Evaluation Results}
Before presenting accuracy metrics outlined in section \ref{subsec:reid_metric}, it is noteworthy to study the computational efficiency of different backbone architectures in terms of throughput, GPU memory utilization, and dimensionality of output features. To design real-time transportation applications at scale, such measures are critical so that their improvement at the expense of slight sacrifice of performance can be reasonably justified. These metrics, reflected in Table \ref{tab:backbone_stats}, are often overlooked in the re-id literature as the main focus is mainly directed towards enhancing accuracy of evaluation metrics.

\begin{table}[t!]
    \caption{Efficiency metrics of different backbones tested on NVIDIA RTX 2080 GPU card. Note that number of parameters, speed, and memory utilization are measured in Millions (M), milliseconds (ms) per image, and MegaBytes (MB) respectively.}
    \centering
    \label{tab:backbone_stats}
    \resizebox{0.8\columnwidth}{!}{
    \begin{tabular}{c|c|c|c|c}
         \hline
         \multicolumn{1}{c|}{Architecture} &\multicolumn{1}{c|}{\# Params (M)} & \multicolumn{1}{c|}{Speed (ms/image)} & \multicolumn{1}{c|}{Memory (MB)} & \multicolumn{1}{|c}{\# Dims} \\ \hline \hline
         \multicolumn{1}{c|}{ResNet50}  &\multicolumn{1}{c|}{$23.51$} & \multicolumn{1}{c|}{$3.22$} & \multicolumn{1}{c|}{$122$} & 
         \multicolumn{1}{|c}{$2048$} \\
         \multicolumn{1}{c|}{ResNet50\_IBN}  &\multicolumn{1}{c|}{$23.51$} & \multicolumn{1}{c|}{$4.55$} & \multicolumn{1}{c|}{$122$} &
         \multicolumn{1}{|c}{$2048$}
         \\
         \multicolumn{1}{c|}{ViT\_Base} &\multicolumn{1}{c|}{$85.50$} & \multicolumn{1}{c|}{$3.80$} & \multicolumn{1}{c|}{$341$} &
         \multicolumn{1}{|c}{$768$}
         \\
         \multicolumn{1}{c|}{SWIN\_Base} &\multicolumn{1}{c|}{$86.74$} & \multicolumn{1}{c|}{$11.94$} & \multicolumn{1}{c|}{$363$} &
         \multicolumn{1}{|c}{$1024$}
         \\
         \multicolumn{1}{c|}{ConvNext\_Base} &\multicolumn{1}{c|}{$87.57$} & \multicolumn{1}{c|}{$6.22$} & \multicolumn{1}{c|}{$355$} &
         \multicolumn{1}{|c}{$1024$}\\
         \hline
    \end{tabular}
    }
\end{table}
From Table \ref{tab:backbone_stats}, we see that ResNet-based models have fewer parameters and memory footprint. But we should note that their learning capacity from large volume of data is not on par with larger models. It is also evident that SWIN\_base has a significantly lower throughput. However, as we will see in section \ref{subsubsec:VeRiWild}, it has the highest performance in the presence of abundant data. We note that there are variants to each particular architecture that should be chosen depending on the use-case and availability of resources.

\subsubsection{Evaluation Results on VeRi Dataset:}
Table \ref{tab:veri_results} reports numbers for both baseline and SSBVER models with different backbone architectures. 
\begin{table}[t]
    \caption{Performance Comparison between SSBVER model against baseline on VeRi datasets. Note that bold black figures denote the higher performance for each architecture while bold red figures are the highest among all models and architectures.}
    \centering
    \label{tab:veri_results}
    \resizebox{0.65\columnwidth}{!}{
    \begin{tabular}{c|c|c|c|c}
         \hline
         \multicolumn{1}{c|}{\multirow{2}{*}{Architecture}} &
         \multicolumn{1}{|c|}{\multirow{2}{*}{Model}} & \multicolumn{3}{||c}{Evaluation Metric} \\
         \cline{3-5} 
         \multicolumn{1}{c|}{} &
         \multicolumn{1}{|c|}{} &
         \multicolumn{1}{||c|}{mAP (\%)} &
         \multicolumn{1}{|c|}{CMC@1 (\%)} &
         \multicolumn{1}{|c}{CMC@5 (\%)} \\
         \hline \hline
         \multicolumn{1}{c|}{\multirow{2}{*}{ResNet50}} & \multicolumn{1}{|c|}{Baseline} & \multicolumn{1}{||c|}{$78.03$} & \multicolumn{1}{|c|}{$95.89$} & \multicolumn{1}{|c}{$97.85$} \\
          \multicolumn{1}{c|}{} & \multicolumn{1}{|c|}{SSBVER} & \multicolumn{1}{||c|}{$\textbf{80.94}$} & \multicolumn{1}{|c|}{$\textbf{97.02}$} & \multicolumn{1}{|c}{$\textbf{98.45}$} \\ \hline \hline
         \multicolumn{1}{c|}{\multirow{2}{*}{ResNet50\_IBN}} & \multicolumn{1}{|c|}{Baseline} & \multicolumn{1}{||c|}{$79.88$} & \multicolumn{1}{|c|}{$96.13$} & \multicolumn{1}{|c}{$97.97$} \\
          \multicolumn{1}{c|}{} & \multicolumn{1}{|c|}{SSBVER} & \multicolumn{1}{||c|}{$\textcolor{red}{\textbf{82.11}}$} & \multicolumn{1}{|c|}{$\textcolor{red}{\textbf{97.08}}$} & \multicolumn{1}{|c}{$\textbf{98.45}$} \\ \hline \hline
         \multicolumn{1}{c|}{\multirow{2}{*}{ViT\_Base}} & \multicolumn{1}{|c|}{Baseline} & \multicolumn{1}{||c|}{$77.72$} & \multicolumn{1}{|c|}{$\textbf{96.48}$} & \multicolumn{1}{|c}{$98.39$} \\
          \multicolumn{1}{c|}{} & \multicolumn{1}{|c|}{SSBVER} & \multicolumn{1}{||c|}{$\textbf{77.74}$} & \multicolumn{1}{|c|}{$95.95$} & \multicolumn{1}{|c}{$\textcolor{red}{\textbf{98.51}}$} \\ \hline \hline
         \multicolumn{1}{c|}{\multirow{2}{*}{SWIN\_Base}} & \multicolumn{1}{|c|}{Baseline} & \multicolumn{1}{||c|}{$78.40$} & \multicolumn{1}{|c|}{$95.65$} & \multicolumn{1}{|c}{$97.85$} \\
          \multicolumn{1}{c|}{} & \multicolumn{1}{|c|}{SSBVER} & \multicolumn{1}{||c|}{$\textbf{79.35}$} & \multicolumn{1}{|c|}{$\textbf{95.74}$} & \multicolumn{1}{|c}{$\textbf{97.91}$} \\ \hline \hline
         \multicolumn{1}{c|}{\multirow{2}{*}{ConvNext\_Base}} & \multicolumn{1}{|c|}{Baseline} & \multicolumn{1}{||c|}{$78.73$} & \multicolumn{1}{|c|}{$96.13$} & \multicolumn{1}{|c}{$98.03$} \\
          \multicolumn{1}{c|}{} & \multicolumn{1}{|c|}{SSBVER} & \multicolumn{1}{||c|}{$\textbf{79.01}$} & \multicolumn{1}{|c|}{$\textbf{96.36}$} & \multicolumn{1}{|c}{$\textbf{98.27}$} \\ \hline
    \end{tabular}
    }
\end{table}
It can be seen that for the VeRi dataset, SSBVER outperforms the baseline model in almost every evaluation metrics and for all backbone architectures with the exception of ViT\_Base model for the CMC@1 metric. We should note that CMC@1 is more sensitive compared to other metrics as it only considers the first item in the ranked gallery which is either a hit or miss. This can also be attributed to the fact that ViT\_Base model has a high capacity for learning in data-abundant regime as noted in \cite{dosovitskiy2020image} which is not the case for VeRi dataset. Therefore, adding self-supervision does not result in learning improved representations. In addition, we highlight that the performance of ResNet50 and ResNet50\_IBN models are significantly improved by adopting self-supervision in training pipeline compared to architectures with larger number of parameters that can easily overfit the data and suffer from high variance. We would like to highlight that the performance gained here is cost-free in that SSLBVER preserves the speed and memory utilization of the baseline model. 
\subsubsection{Evaluation Results on VehicleID Dataset:}
\begin{table}[]
    \caption{Performance Comparison between SSBVER model against baseline on VehicleID datasets. Note that bold black figures denote the higher performance for each architecture while bold red figures are the highest among all models and architectures.}
    \centering
    \label{tab:vehicleid_results}
    \resizebox{.9\columnwidth}{!}{
    \begin{tabular}{c|c|c|c|c|c|c|c|c|c|c}
         \hline
         \multicolumn{1}{c|}{\multirow{3}{*}{Architecture}} &
         \multicolumn{1}{|c|}{\multirow{3}{*}{Model}} & \multicolumn{9}{||c}{Evaluation Metric} \\
         \cline{3-11} 
         \multicolumn{1}{c|}{} &
         \multicolumn{1}{|c|}{} &
         \multicolumn{3}{||c|}{mAP (\%)} &
         \multicolumn{3}{||c|}{CMC@1 (\%)} &
         \multicolumn{3}{||c}{CMC@5 (\%)} \\
         \cline{3-11}
         \multicolumn{1}{c|}{} &
         \multicolumn{1}{|c|}{} &
         \multicolumn{1}{||c|}{S} &
         \multicolumn{1}{|c|}{M} &
         \multicolumn{1}{|c}{L} &
         \multicolumn{1}{||c|}{S} &
         \multicolumn{1}{|c|}{M} &
         \multicolumn{1}{|c}{L} &
         \multicolumn{1}{||c|}{S} &
         \multicolumn{1}{|c|}{M} &
         \multicolumn{1}{|c}{L} \\
         \hline \hline
         \multicolumn{1}{c|}{\multirow{2}{*}{ResNet50}} & \multicolumn{1}{|c|}{Baseline} & \multicolumn{1}{||c|}{$88.77$} & \multicolumn{1}{|c|}{$86.05$} & 
         \multicolumn{1}{|c}{$82.91$} & \multicolumn{1}{||c|}{$82.75$} & \multicolumn{1}{|c|}{$80.26$} & 
         \multicolumn{1}{|c}{$76.79$} & \multicolumn{1}{||c|}{$97.00$} & \multicolumn{1}{|c|}{$94.06$} & 
         \multicolumn{1}{|c}{$90.92$} 
         \\
          \multicolumn{1}{c|}{} & 
          \multicolumn{1}{|c|}{SSBVER} & \multicolumn{1}{||c|}{$\textbf{90.73}$} & \multicolumn{1}{|c|}{$\textbf{86.57}$} & \multicolumn{1}{|c}{$\textbf{83.82}$} & \multicolumn{1}{||c|}{$\textbf{85.61}$} & \multicolumn{1}{|c|}{$\textbf{80.34}$} & \multicolumn{1}{|c}{$\textbf{77.26}$} & \multicolumn{1}{||c|}{$\textbf{97.73}$} & \multicolumn{1}{|c|}{$\textbf{94.92}$} & \multicolumn{1}{|c}{$\textbf{92.59}$}
          \\ \hline \hline
         \multicolumn{1}{c|}{\multirow{2}{*}{ResNet50\_IBN}} & \multicolumn{1}{|c|}{Baseline} & \multicolumn{1}{||c|}{$89.19$} & \multicolumn{1}{|c|}{$84.95$} & 
         \multicolumn{1}{|c}{$82.73$} & \multicolumn{1}{||c|}{$83.44$} & \multicolumn{1}{|c|}{$78.81$} & 
         \multicolumn{1}{|c}{$76.79$} & \multicolumn{1}{||c|}{$96.82$} & \multicolumn{1}{|c|}{$93.13$} & 
         \multicolumn{1}{|c}{$90.53$} 
         \\
          \multicolumn{1}{c|}{} & 
          \multicolumn{1}{|c|}{SSBVER} & \multicolumn{1}{||c|}{$\textcolor{red}{\textbf{90.88}}$} & 
          \multicolumn{1}{|c|}{$\textcolor{red}{\textbf{87.36}}$} & 
          \multicolumn{1}{|c}{$\textcolor{red}{\textbf{84.83}}$} & \multicolumn{1}{||c|}{$\textcolor{red}{\textbf{85.61}}$} & 
          \multicolumn{1}{|c|}{$\textcolor{red}{\textbf{81.62}}$} & 
          \multicolumn{1}{|c}{$\textcolor{red}{\textbf{78.91}}$} & \multicolumn{1}{||c|}{$\textbf{97.72}$} & 
          \multicolumn{1}{|c|}{$\textbf{94.92}$} & 
          \multicolumn{1}{|c}{$\textbf{92.60}$}
          \\ \hline \hline
         \multicolumn{1}{c|}{\multirow{2}{*}{ViT\_Base}} & \multicolumn{1}{|c|}{Baseline} & \multicolumn{1}{||c|}{$88.70$} & \multicolumn{1}{|c|}{$84.88$} & 
         \multicolumn{1}{|c}{$82.65$} & \multicolumn{1}{||c|}{$82.50$} & \multicolumn{1}{|c|}{$78.53$} & 
         \multicolumn{1}{|c}{$76.33$} & \multicolumn{1}{||c|}{$97.22$} & \multicolumn{1}{|c|}{$\textbf{93.61}$} 
         & \multicolumn{1}{|c}{$90.75$}\\
          \multicolumn{1}{c|}{} & \multicolumn{1}{|c|}{SSBVER} & \multicolumn{1}{||c|}{$\textbf{89.09}$} & \multicolumn{1}{|c|}{$\textbf{85.23}$} & \multicolumn{1}{|c}{$\textbf{83.13}$} & \multicolumn{1}{||c|}{$\textbf{82.93}$} & \multicolumn{1}{|c|}{$\textbf{79.05}$} & \multicolumn{1}{|c}{$\textbf{76.64}$} & \multicolumn{1}{||c|}{$\textbf{97.33}$} & \multicolumn{1}{|c|}{$93.56$} & \multicolumn{1}{|c}{$\textbf{91.78}$} 
          \\ \hline \hline
         \multicolumn{1}{c|}{\multirow{2}{*}{SWIN\_Base}} & \multicolumn{1}{|c|}{Baseline} & \multicolumn{1}{||c|}{$89.77$} & \multicolumn{1}{|c|}{$86.74$} & 
         \multicolumn{1}{|c}{$84.35$} & \multicolumn{1}{||c|}{$83.84$} & \multicolumn{1}{|c|}{$80.86$} & 
         \multicolumn{1}{|c}{$77.90$} & \multicolumn{1}{||c|}{$97.61$} & \multicolumn{1}{|c|}{$94.91$} & 
         \multicolumn{1}{|c}{$92.17$}
         \\
          \multicolumn{1}{c|}{} & 
          \multicolumn{1}{|c|}{SSBVER} & \multicolumn{1}{||c|}{$\textbf{90.58}$} & \multicolumn{1}{|c|}{$\textbf{86.98}$} & \multicolumn{1}{|c}{$\textbf{84.68}$} & \multicolumn{1}{||c|}{$\textbf{85.19}$} & \multicolumn{1}{|c|}{$\textbf{81.02}$} & \multicolumn{1}{|c}{$\textbf{78.62}$} & \multicolumn{1}{||c|}{$\textcolor{red}{\textbf{97.96}}$} & \multicolumn{1}{|c|}{$\textcolor{red}{\textbf{95.08}}$} & \multicolumn{1}{|c}{$\textcolor{red}{\textbf{93.27}}$} \\ \hline \hline
         \multicolumn{1}{c|}{\multirow{2}{*}{ConvNext\_Base}} & \multicolumn{1}{|c|}{Baseline} & \multicolumn{1}{||c|}{$88.95$} & \multicolumn{1}{|c|}{$85.54$} & \multicolumn{1}{|c}{$83.14$} & \multicolumn{1}{||c|}{$82.84$} & \multicolumn{1}{|c|}{$\textbf{79.46}$} & \multicolumn{1}{|c}{$76.69$} & \multicolumn{1}{||c|}{$97.03$} & \multicolumn{1}{|c|}{$93.69$} & \multicolumn{1}{|c}{$\textbf{91.51}$}\\
          \multicolumn{1}{c|}{} & \multicolumn{1}{|c|}{SSBVER} & \multicolumn{1}{||c|}{$\textbf{89.10}$} & \multicolumn{1}{|c|}{$\textbf{85.81}$} & \multicolumn{1}{|c}{$\textbf{83.24}$} & \multicolumn{1}{||c|}{$\textbf{83.42}$} & \multicolumn{1}{|c|}{$79.38$} & \multicolumn{1}{|c}{$\textbf{77.13}$} & \multicolumn{1}{||c|}{$\textbf{97.17}$} & \multicolumn{1}{|c|}{$\textbf{94.17}$} & \multicolumn{1}{|c}{$91.44$} \\ \hline
    \end{tabular}
    }
\end{table}
As mentioned earlier, the test set of VehicleID dataset has three splits: small, medium and large which are enumerated by S, M, and L in Table \ref{tab:vehicleid_results} respectively. We should emphasize that images in VehicleID are only captured from either front or rear and the extent to which a network can exploit small-scale information in overlapping views,  which is critical for re-id, is limited. Similar to VeRi dataset, self-supervised objective contributes to performance improvement across all evaluation metrics and for all backbone architectures. Due to its relatively larger size compared to VeRi dataset, the performance of bigger models, namely ViT, SWIN and ConvNext is much closer to ResNet50 and ResNet50\_IBN. In particular, for this dataset ResNet50\_IBN, and SWIN\_Base architectures mostly yield highest scores. The superior performance of SWIN compared to ViT shows the benefit of hierarchical design and multi-resolution feature maps in a transformer-based model as it can better extract information at various scales. 

\subsubsection{Evaluation Results on VeRiWild Dataset:}
\label{subsubsec:VeRiWild}
Similar to the VehicleID dataset, the test set of VeRiWild is split into three small, medium and large sets consisting of $41861$, $69389$, and $138517$ images respectively. The performance of the baseline and SSBVER models with different architectures are reported in Table \ref{tab:veriwild_results}. For this multi-view dataset the benefit of self-supervision in the form of knowledge distillation and self-distillation is quite evident as every evaluation metric across all the test splits and architectures is improved by a significant margin. This shows that DINO objective effectively regulates the model training to exploit more fine-grained features that are favorable for the vehicle re-id task. Moreover, because of the large number of training samples, ViT, SWIN, and Convnext achieve substantially higher performance compared to ResNet-based models. Here SWIN gives the highest performance despite having the lowest throughput. We note that SWIN achieves the highest performance on VeRiWild compared to all the state-of-the-art models. 
\begin{table}[]
    \caption{Performance Comparison between SSBVER model against baseline on VeRiWild datasets. Note that bold black figures denote the higher performance for each architecture while bold red figures are the highest among all models and architectures.}
    \centering
    \label{tab:veriwild_results}
    \resizebox{.9\columnwidth}{!}{
    \begin{tabular}{c|c|c|c|c|c|c|c|c|c|c}
         \hline
         \multicolumn{1}{c|}{\multirow{3}{*}{Architecture}} &
         \multicolumn{1}{|c|}{\multirow{3}{*}{Model}} & \multicolumn{9}{||c}{Evaluation Metric} \\
         \cline{3-11} 
         \multicolumn{1}{c|}{} &
         \multicolumn{1}{|c|}{} &
         \multicolumn{3}{||c|}{mAP (\%)} &
         \multicolumn{3}{||c|}{CMC@1 (\%)} &
         \multicolumn{3}{||c}{CMC@5 (\%)} \\
         \cline{3-11}
         \multicolumn{1}{c|}{} &
         \multicolumn{1}{|c|}{} &
         \multicolumn{1}{||c|}{S} &
         \multicolumn{1}{|c|}{M} &
         \multicolumn{1}{|c}{L} &
         \multicolumn{1}{||c|}{S} &
         \multicolumn{1}{|c|}{M} &
         \multicolumn{1}{|c}{L} &
         \multicolumn{1}{||c|}{S} &
         \multicolumn{1}{|c|}{M} &
         \multicolumn{1}{|c}{L} \\
         \hline \hline
         \multicolumn{1}{c|}{\multirow{2}{*}{ResNet50}} & \multicolumn{1}{|c|}{Baseline} & 
         \multicolumn{1}{||c|}{$78.20$} &
         \multicolumn{1}{|c|}{$72.43$} & 
         \multicolumn{1}{|c}{$64.43$} & 
         \multicolumn{1}{||c|}{$93.14$} & 
         \multicolumn{1}{|c|}{$90.62$} & 
         \multicolumn{1}{|c}{$86.93$} & 
         \multicolumn{1}{||c|}{$97.82$} & 
         \multicolumn{1}{|c|}{$96.89$} & 
         \multicolumn{1}{|c}{$94.71$} 
         \\
          \multicolumn{1}{c|}{} & 
          \multicolumn{1}{|c|}{SSBVER} &
          \multicolumn{1}{||c|}{$\textbf{80.41}$} &
          \multicolumn{1}{|c|}{$\textbf{74.77}$} &
          \multicolumn{1}{|c}{$\textbf{67.02}$} &
          \multicolumn{1}{||c|}{$\textbf{93.88}$} &
          \multicolumn{1}{|c|}{$\textbf{91.44}$} &
          \multicolumn{1}{|c}{$\textbf{88.26}$} &
          \multicolumn{1}{||c|}{$\textbf{98.03}$} &
          \multicolumn{1}{|c|}{$\textbf{96.93}$} &
          \multicolumn{1}{|c}{$\textbf{94.98}$}
          \\ \hline \hline
         \multicolumn{1}{c|}{\multirow{2}{*}{ResNet50\_IBN}} & \multicolumn{1}{|c|}{Baseline} &
         \multicolumn{1}{||c|}{$81.46$} & 
         \multicolumn{1}{|c|}{$75.74$} & 
         \multicolumn{1}{|c}{$67.70$} & 
         \multicolumn{1}{||c|}{$93.24$} & 
         \multicolumn{1}{|c|}{$90.76$} & 
         \multicolumn{1}{|c}{$86.41$} & 
         \multicolumn{1}{||c|}{$97.82$} & 
         \multicolumn{1}{|c|}{$96.51$} & 
         \multicolumn{1}{|c}{$94.20$} 
         \\
          \multicolumn{1}{c|}{} & 
          \multicolumn{1}{|c|}{SSBVER} &
          \multicolumn{1}{||c|}{$\textbf{82.64}$} & 
          \multicolumn{1}{|c|}{$\textbf{77.49}$} & 
          \multicolumn{1}{|c}{$\textbf{70.09}$} &
          \multicolumn{1}{||c|}{$\textbf{95.11}$} & 
          \multicolumn{1}{|c|}{$\textbf{93.37}$} & 
          \multicolumn{1}{|c}{$\textbf{90.14}$} &
          \multicolumn{1}{||c|}{$\textbf{98.53}$} & 
          \multicolumn{1}{|c|}{$\textbf{97.45}$} & 
          \multicolumn{1}{|c}{$\textbf{95.67}$}
          \\ \hline \hline
         \multicolumn{1}{c|}{\multirow{2}{*}{ViT\_Base}} & \multicolumn{1}{|c|}{Baseline} & \multicolumn{1}{||c|}{$81.76$} & 
         \multicolumn{1}{|c|}{$76.13$} & 
         \multicolumn{1}{|c}{$67.71$} & 
         \multicolumn{1}{||c|}{$93.44$} & 
         \multicolumn{1}{|c|}{$91.56$} & 
         \multicolumn{1}{|c}{$86.77$} & 
         \multicolumn{1}{||c|}{$98.59$} & 
         \multicolumn{1}{|c|}{$97.57$} & 
         \multicolumn{1}{|c}{$95.55$}\\
          \multicolumn{1}{c|}{} & 
          \multicolumn{1}{|c|}{SSBVER} & \multicolumn{1}{||c|}{$\textbf{83.81}$} & \multicolumn{1}{|c|}{$\textbf{78.25}$} &
          \multicolumn{1}{|c}{$\textbf{70.55}$} &
          \multicolumn{1}{||c|}{$\textbf{94.98}$} &
          \multicolumn{1}{|c|}{$\textbf{92.71}$} &
          \multicolumn{1}{|c}{$\textbf{89.65}$} &
          \multicolumn{1}{||c|}{$\textbf{98.69}$} &
          \multicolumn{1}{|c|}{$\textbf{97.83}$} &
          \multicolumn{1}{|c}{$\textbf{95.98}$} 
          \\ \hline \hline
         \multicolumn{1}{c|}{\multirow{2}{*}{SWIN\_Base}} & \multicolumn{1}{|c|}{Baseline} & \multicolumn{1}{||c|}{$84.94$} & 
         \multicolumn{1}{|c|}{$79.64$} & 
         \multicolumn{1}{|c}{$71.93$} & 
         \multicolumn{1}{||c|}{$94.58$} & 
         \multicolumn{1}{|c|}{$92.05$} & 
         \multicolumn{1}{|c}{$87.89$} & 
         \multicolumn{1}{||c|}{$98.80$} & 
         \multicolumn{1}{|c|}{$97.55$} & 
         \multicolumn{1}{|c}{$95.97$}
         \\
          \multicolumn{1}{c|}{} & 
          \multicolumn{1}{|c|}{SSBVER} &
          \multicolumn{1}{||c|}{$\textcolor{red}{\textbf{86.05}}$} &
          \multicolumn{1}{|c|}{$\textcolor{red}{\textbf{81.28}}$} & \multicolumn{1}{|c}{$\textcolor{red}{\textbf{74.07}}$} & \multicolumn{1}{||c|}{$\textcolor{red}{\textbf{95.62}}$} &
          \multicolumn{1}{|c|}{$\textcolor{red}{\textbf{93.75}}$} & \multicolumn{1}{|c}{$\textcolor{red}{\textbf{90.27}}$} & \multicolumn{1}{||c|}{$\textcolor{red}{\textbf{99.10}}$} &
          \multicolumn{1}{|c|}{$\textcolor{red}{\textbf{98.23}}$} & \multicolumn{1}{|c}{$\textcolor{red}{\textbf{96.76}}$} \\ \hline \hline
         \multicolumn{1}{c|}{\multirow{2}{*}{ConvNext\_Base}} &
         \multicolumn{1}{|c|}{Baseline} &
         \multicolumn{1}{||c|}{$83.44$} & 
         \multicolumn{1}{|c|}{$78.12$} &
         \multicolumn{1}{|c}{$69.93$} &
         \multicolumn{1}{||c|}{$93.74$} & 
         \multicolumn{1}{|c|}{$91.32$} &
         \multicolumn{1}{|c}{$86.69$} &
         \multicolumn{1}{||c|}{$98.33$} & 
         \multicolumn{1}{|c|}{$97.61$} &
         \multicolumn{1}{|c}{$95.59$}\\
          \multicolumn{1}{c|}{} & 
          \multicolumn{1}{|c|}{SSBVER} &
          \multicolumn{1}{||c|}{$\textbf{84.34}$} &
          \multicolumn{1}{|c|}{$\textbf{79.08}$} &
          \multicolumn{1}{|c}{$\textbf{71.29}$} &
          \multicolumn{1}{||c|}{$\textbf{94.21}$} &
          \multicolumn{1}{|c|}{$\textbf{92.29}$} &
          \multicolumn{1}{|c}{$\textbf{88.14}$} &
          \multicolumn{1}{||c|}{$\textbf{98.76}$} &
          \multicolumn{1}{|c|}{$\textbf{97.75}$} &
          \multicolumn{1}{|c}{$\textbf{96.10}$} \\ \hline
    \end{tabular}
    }
\end{table}

\subsubsection{Intra-class Compactness and Inter-class Separation:}
\label{subsec:intra-class-compcatness}
\begin{figure}[b!]
    \centering
    \subfloat[][{\scriptsize VeRi\_Baseline}]{\includegraphics[width=.3\textwidth]{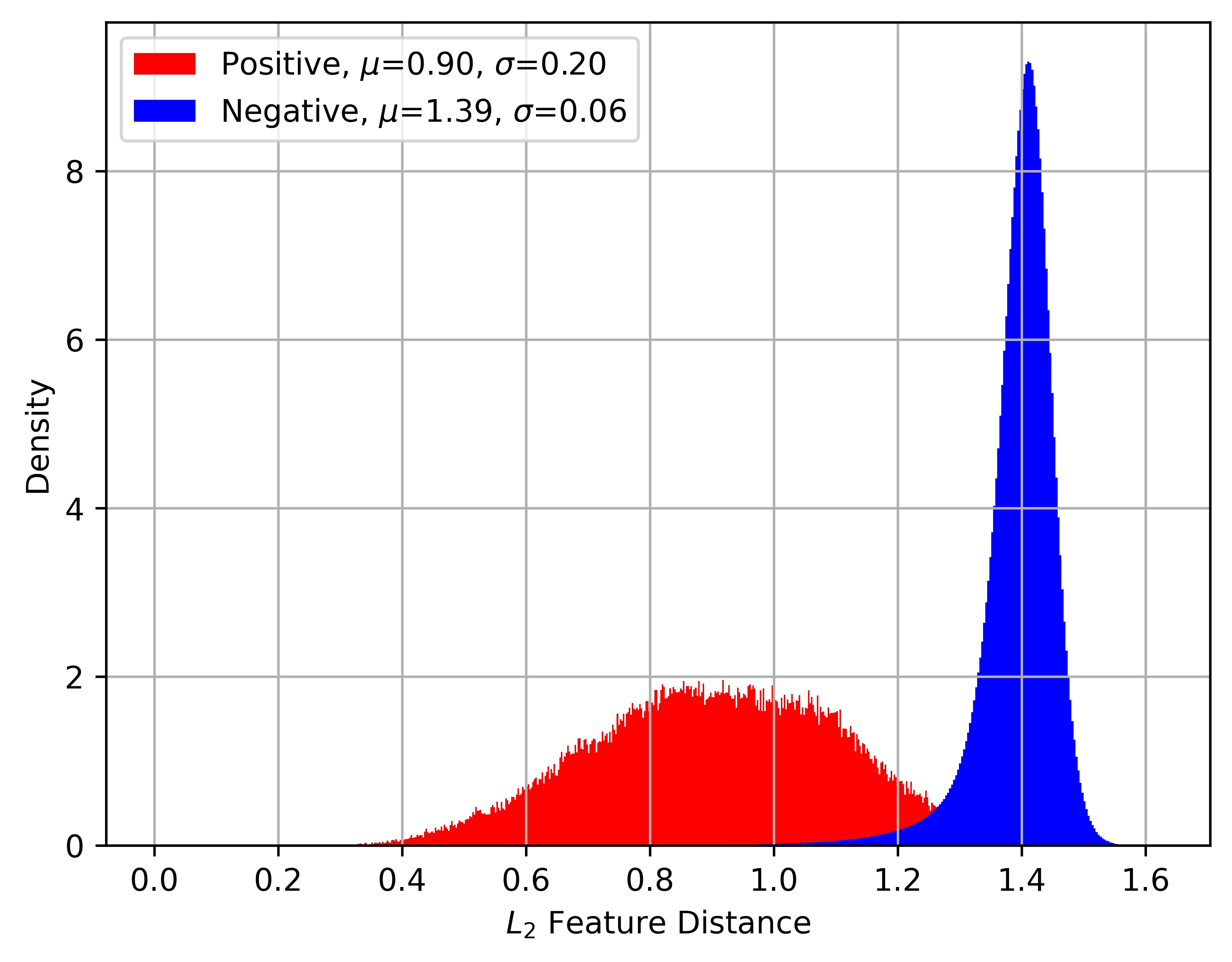}}~
    \subfloat[][{\scriptsize VehicleID\_Baseline }]{\includegraphics[width=0.3\textwidth]{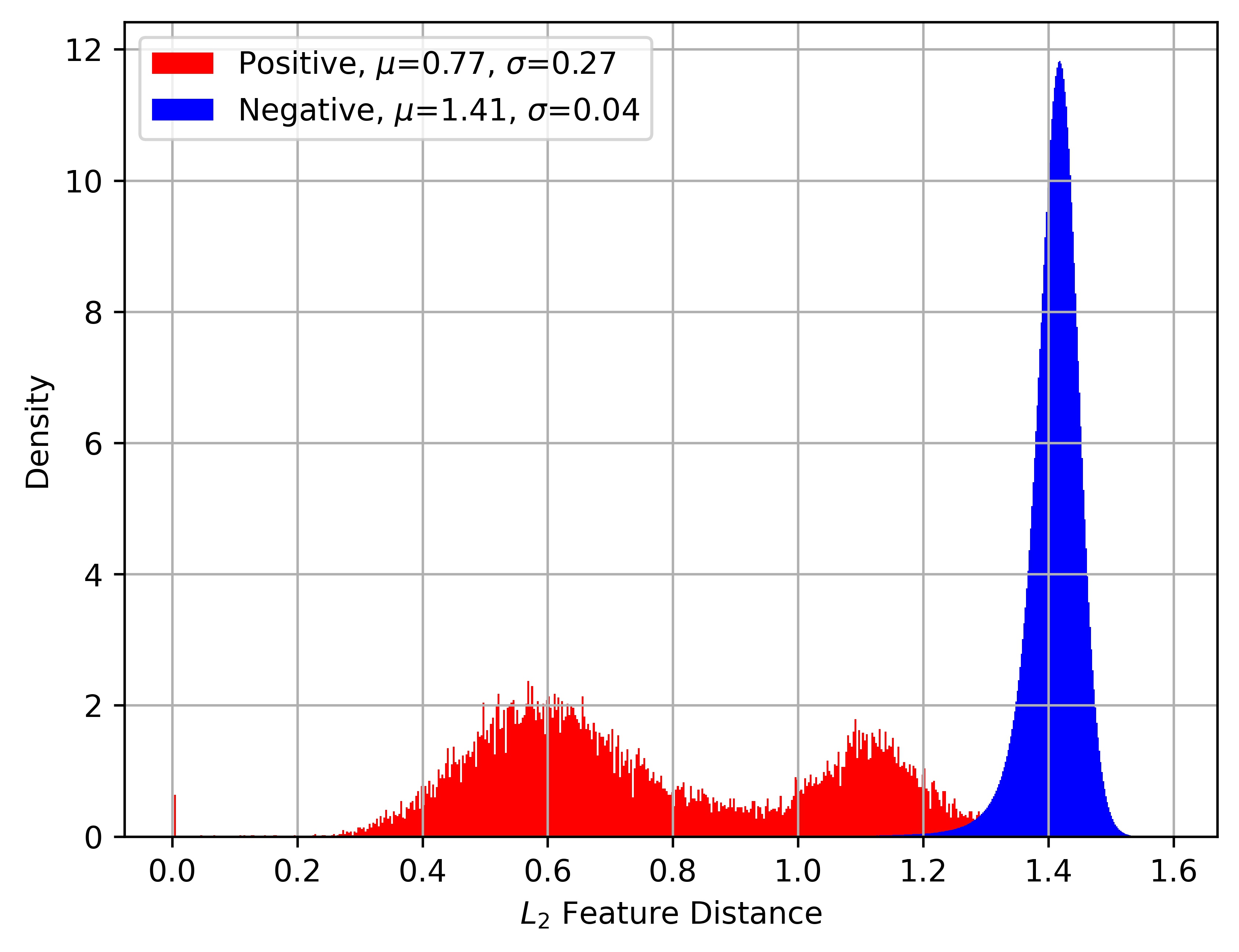}}~
    \subfloat[][{\scriptsize VeRiWild\_Baseline }]{\includegraphics[width=0.3\textwidth]{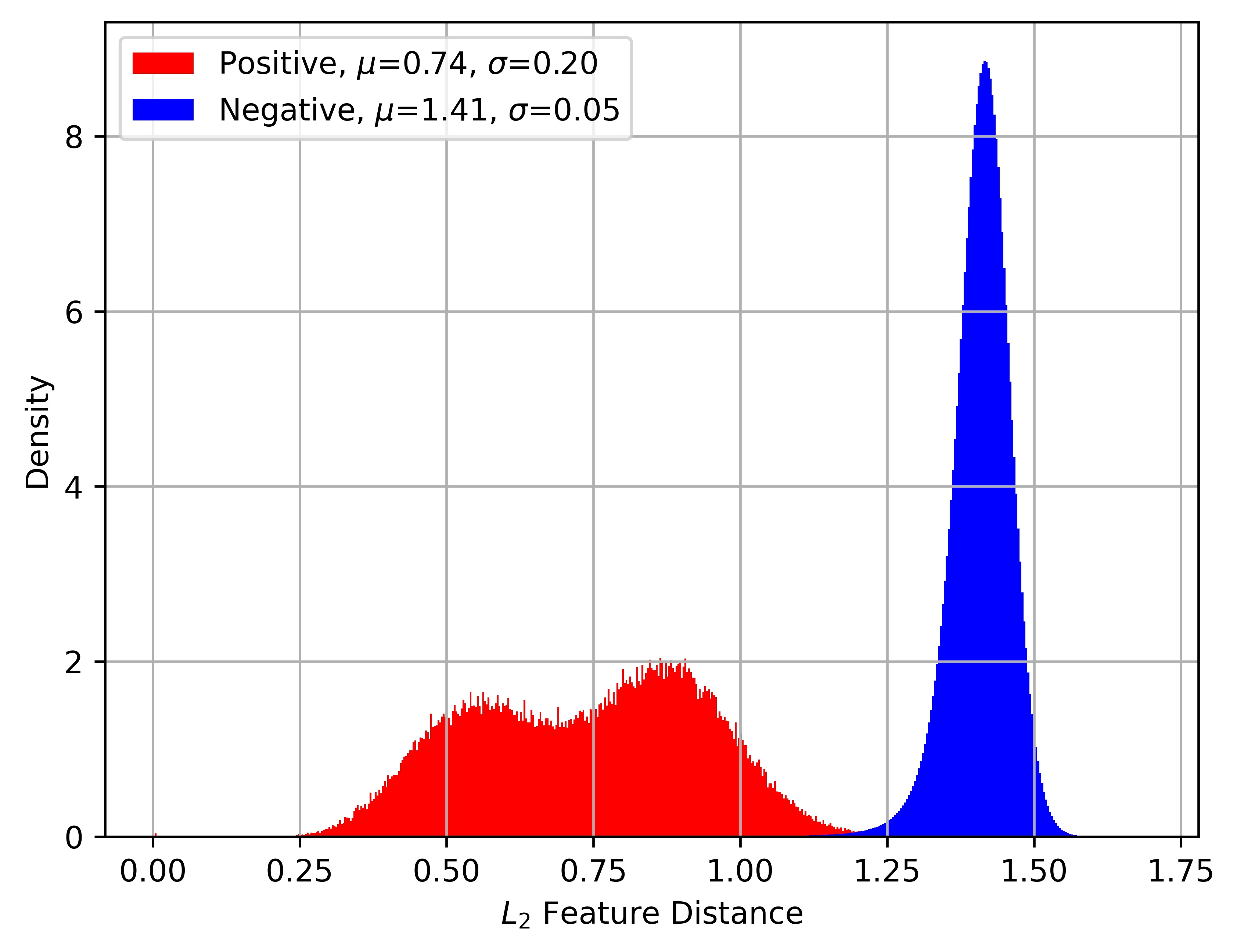}}\\
    \subfloat[][{\scriptsize VeRi\_SSBVER}]{\includegraphics[width=0.3\textwidth]{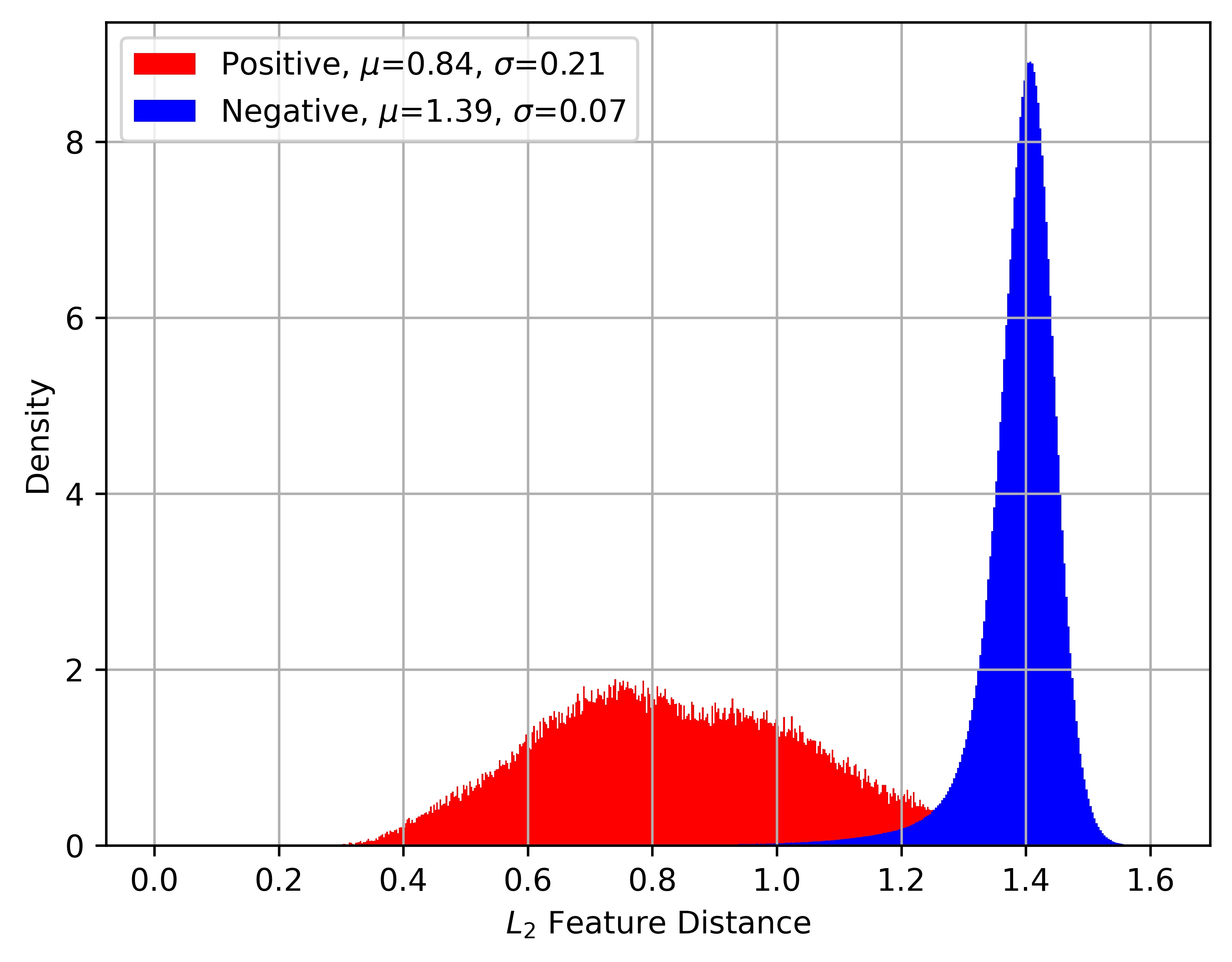}}~
    \subfloat[][{\scriptsize VehicleID\_SSBVER }]{\includegraphics[width=0.3\textwidth]{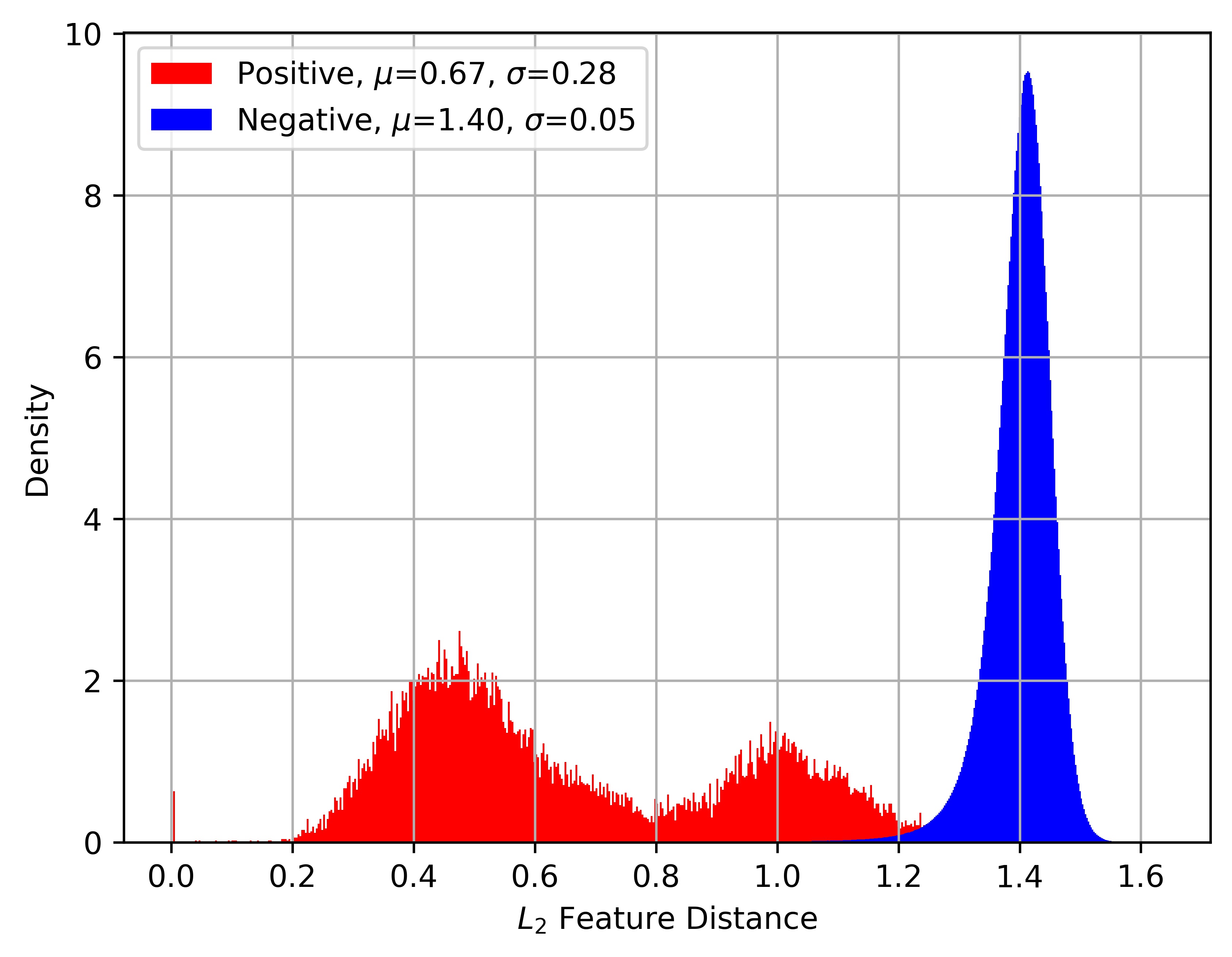}}~
    \subfloat[][{\scriptsize VeRiWild\_SSBVER }]{\includegraphics[width=0.3\textwidth]{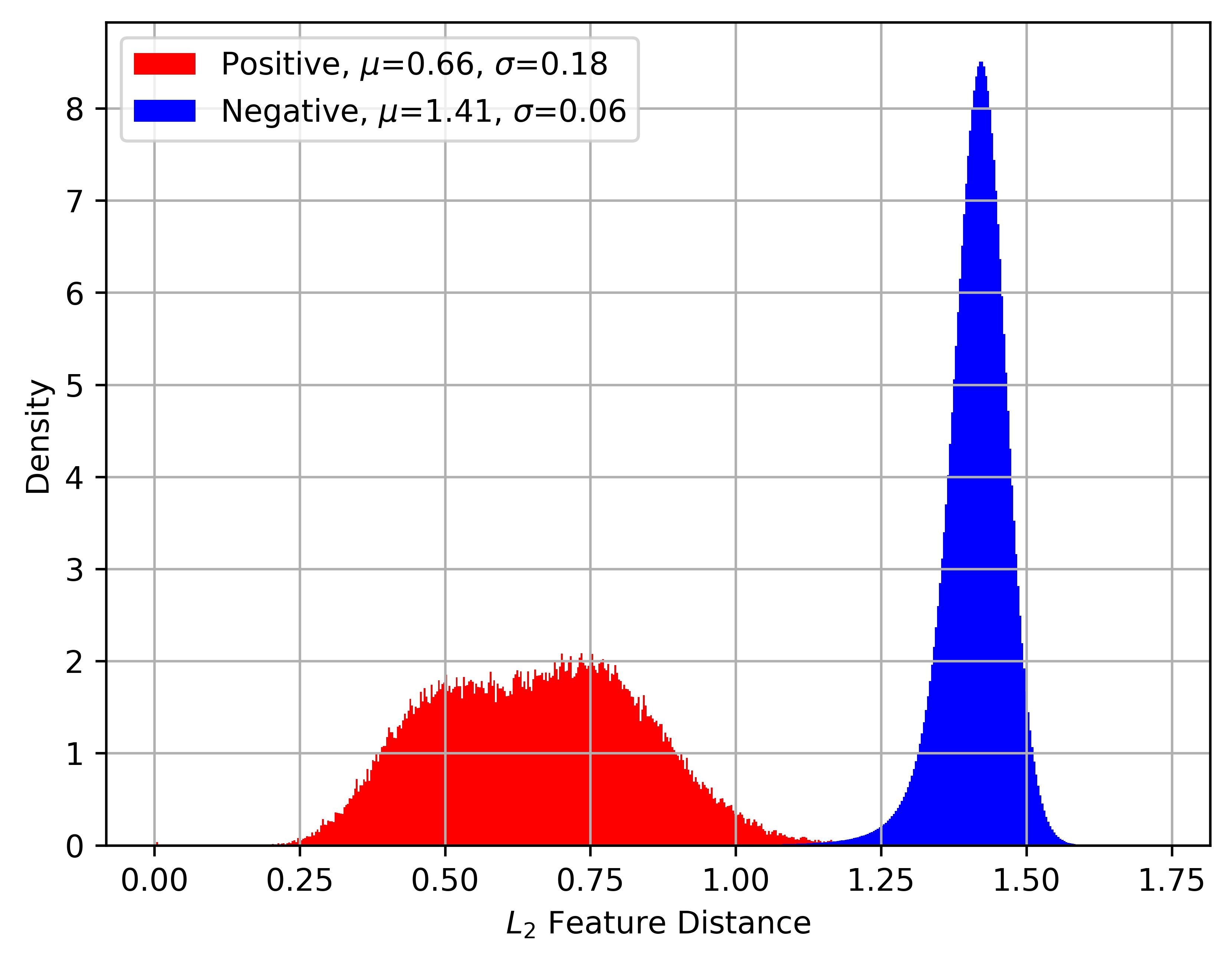}}
    \caption{Distribution of the normalized embedding's distance for positive and negative pairs. The embeddings are extracted from the ResNet50\_IBN backbone architecture. Images on the top row correspond to the baseline configuration and images at the bottom row are based on SSBVER setup.}
    \label{fig:feat_distant_dist}
\end{figure}
Here we are interested in understanding how the incorporation of self-supervised training impacts the $L_2$ distance between extracted features. To this end, we plot the distribution of Euclidean distance of extracted representations for positive (same identity) and negative (different identities) image pairs. Qualitatively, this shows the intra-class compactness and inter-class separation. Fig. \ref{fig:feat_distant_dist} shows the comparison of distance distribution for positive and negative pairs between the baseline and self-supervised boosted models across the test set of different re-id benchmarks. It is seen that DINO objective reduces the mean $\mu$ of positive pair distance distribution by $0.06$, $0.10$, and $0.08$ for VeRi, VehicleID, and VeRiWild datasets respectively. However, the mean of negative pair distance distribution is roughly unchanged. This analysis shows that the incorporation of self-supervision helps the intra-class compactness since the student model is constrained to match the predictions of the teacher model for different views of a same object. Interestingly, for the case of VehicleID, there are two prominent peaks in the distribution of positive pair distances. As images are either captured from the rear or front of vehicles, for positive pairs, the distance is small when both images are from the same views and it is larger when they are from opposing views. Therefore two peaks stand out in the corresponding distribution.

\subsubsection{Convergence Analysis:}
\begin{figure}[t!]
    \centering
    \subfloat[][{\scriptsize Classification Loss}]{\includegraphics[width=.33\textwidth]{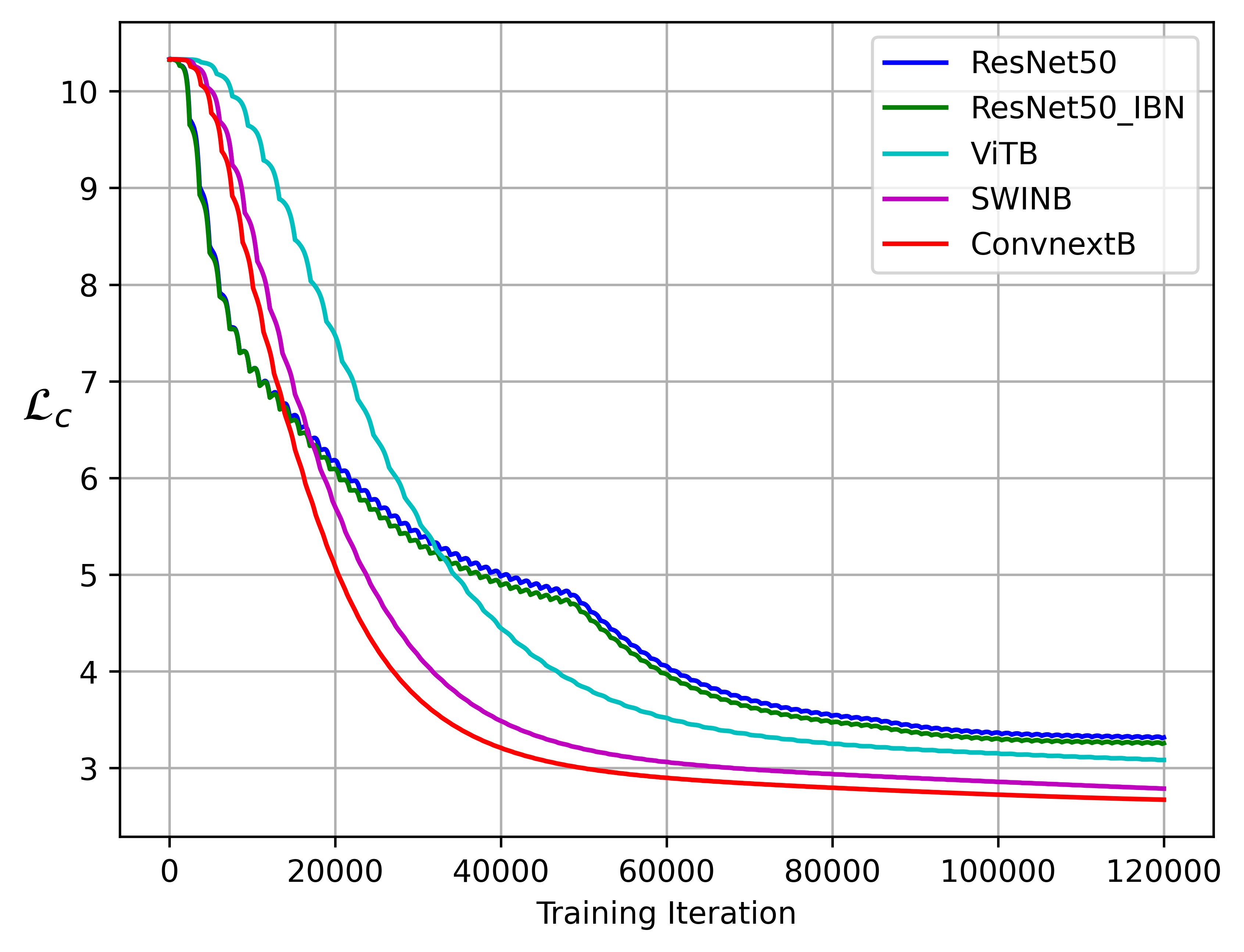}}~
    \subfloat[][{\scriptsize Triplet Loss }]{\includegraphics[width=0.33\textwidth]{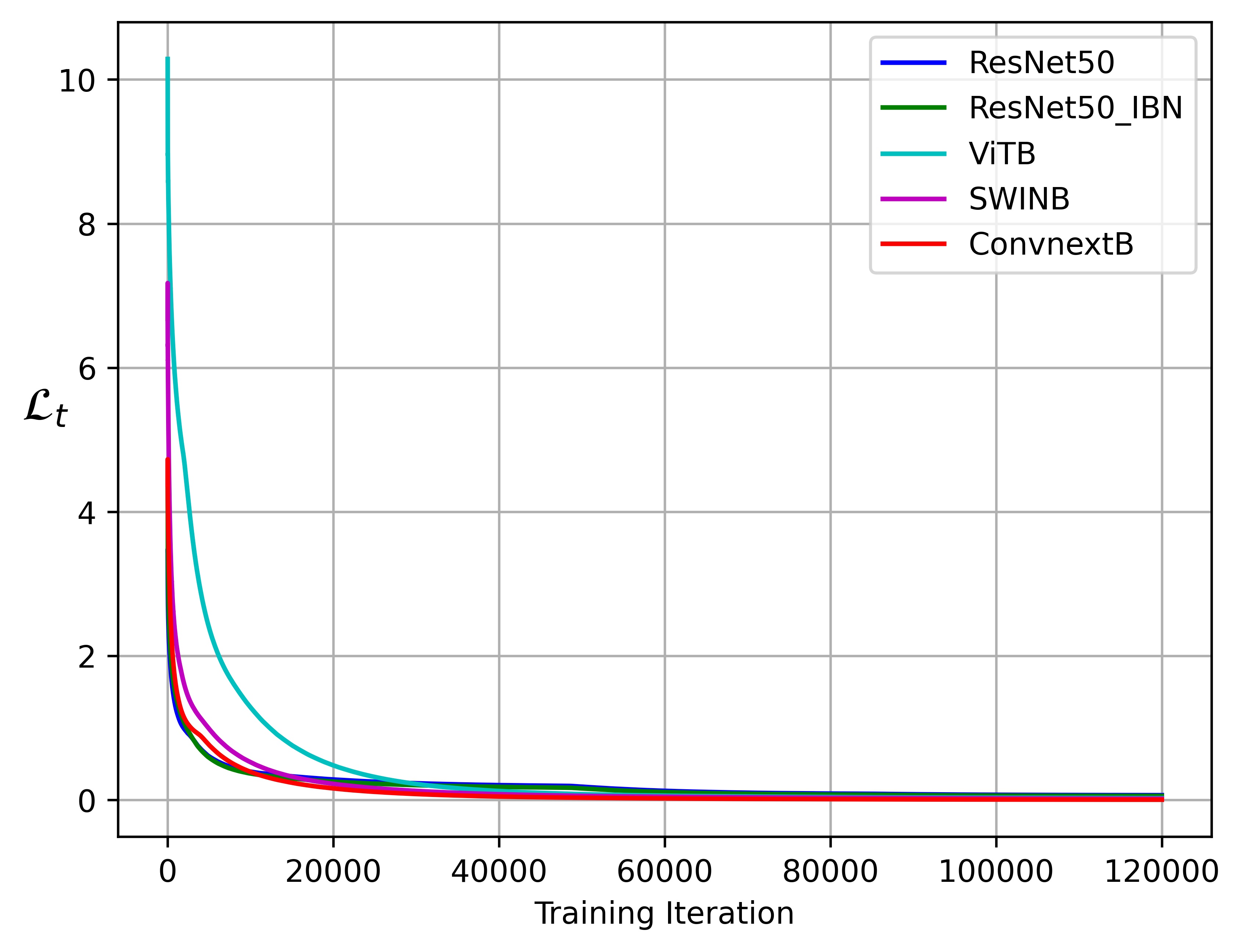}}~
    \subfloat[][{\scriptsize Self-supervised Loss }]{\includegraphics[width=0.33\textwidth]{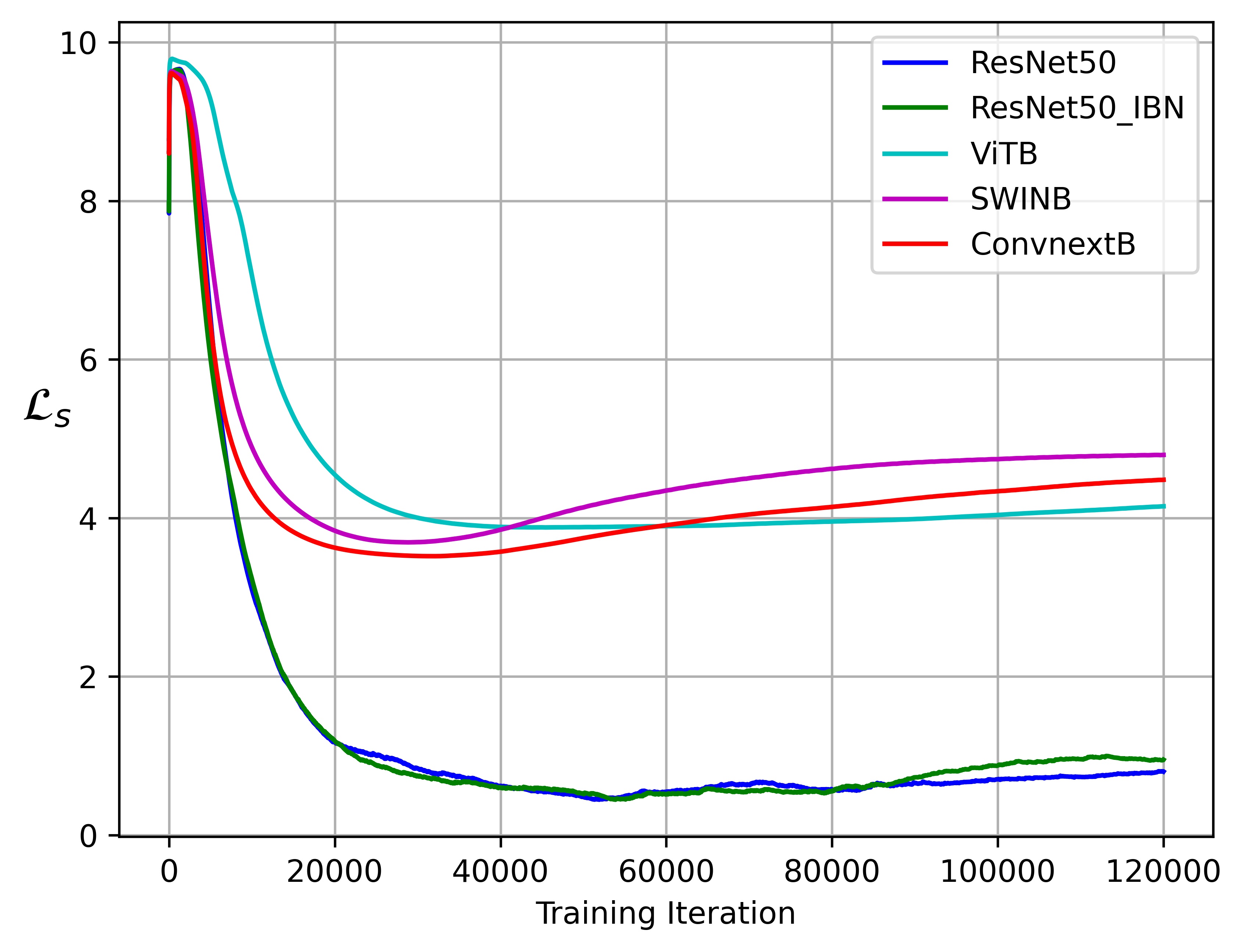}}\\
    \caption{Progression of different objective functions involved in the SSBVER pipeline over the course of training for VeRiWild dataset. Note the exponential moving average is used for each curve to smooth out the fluctuations.}
    \label{fig:Loss}
\end{figure}
In this section we examine the training convergence of the proposed vehicle re-id SSBVER. To this end, we plot the classification, triplet, and self-supervised loss functions during the course of training on VeRiWild dataset in Fig. \ref{fig:Loss}. All the objective functions converge for all the backbone architectures. Since we use label smoothing for the classification objective, its value does not converge to zero. However, Convnext, SWIN, and ViT achieve a lower classification objective compared to ResNet50 and ResNet50\_IBN due to their higher capacity to fit the data. The triplet loss $\mathcal{L}_t$ converges to zero for all the models; although, the rate of convergence for ViT architecture is lower and over the initial training iterations the maximum $L_2$ distance between the features of positive pairs is significantly larger than the minimum $L_2$ distance between the features for negatives resulting in a higher objective value. This can be attributed to the fact that unlike ResNet, SWIN, and Convnext, ViT does not have a hierarchical design and instead has a global receptive field from the first layer. Self-supervised objective evolves differently for ViT, SWIN, and ConvNext compared to ResNet50, and ResNet50\_IBN architectures. While collapse is avoided and the $\mathcal{L}_s$ converges for both groups, it converges to a much lower value for ResNet-based models. This difference can be justified based on the fact that ViT, SWIN, and ConvNext models all use the patchification strategy in their initial layer compared to the down-sampling in ResNet-based models. Down-sampling can provide a better chance for the student model to match the teacher's prediction and achieve a lower objective value.

\subsubsection{Self-supervision as an Implicit Attention Mechanism:}
\begin{figure}[t!]
    \centering
    \subfloat[][{\scriptsize Query\_Baseline}]{\includegraphics[width=.25\textwidth]{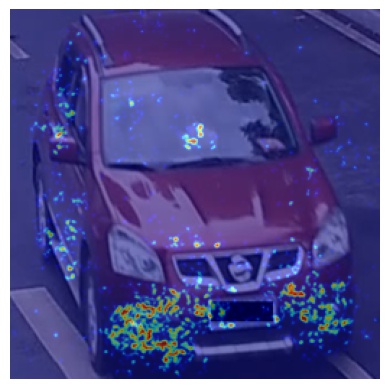}}~\hspace{-.1cm}
    \subfloat[][{\scriptsize Gallery\_Baseline}]{\includegraphics[width=0.25\textwidth]{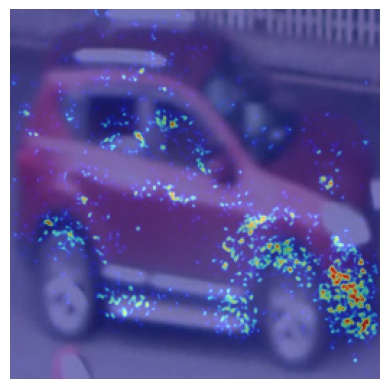}}~\hspace{-.1cm}
    \subfloat[][{\scriptsize Query\_SSBVER}]{\includegraphics[width=0.25\textwidth]{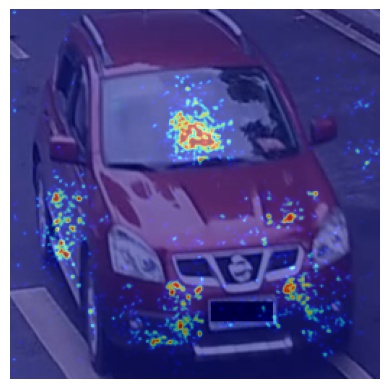}}~\hspace{-.1cm}
    \subfloat[][{\scriptsize Gallery\_SSBVER}]{\includegraphics[width=0.25\textwidth]{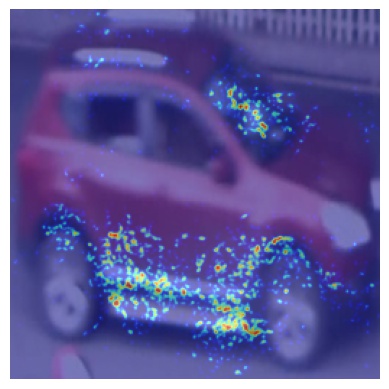}}
    \caption{Saliency maps obtained from baseline and SSBVER models for a pair of query-gallery images. The similarity score generated from baseline model is $0.96$ while SSBVER model yields score of $0.98$. Note that the adopted backbone architecture is ResNet50\_IBN and images are chosen from VeRi dataset.}
    \label{fig:saliency}
\end{figure}
Finally, we visualize the regions of a query and gallery image pair that are most impactful to the similarity score obtained for the pair using SSBVER. More precisely, we compute the input saliency maps for query $m_q$ and gallery $m_g$ via computing the gradient of similarity score for extracted feature vectors with respect to the input images:
\begin{equation}
    m_q = {\nabla}_{I_q} {f_t(I_q)}.{f_t(I_g)}, \quad m_g = {\nabla}_{I_g} {f_t(I_q)}.{f_t(I_g)}
\end{equation}
where $.$ is the dot product and $I_q$, $I_g$ represent query and gallery images respectively. Note that once training is finished, the teacher model is used for testing, therefore, to compute embeddings teacher model $f_t$ is used. Here we did not use the Gradient Class Activation Map (Grad-CAM) \cite{selvaraju2017grad} method as it computes gradient maps in much lower resolutions which results in blob-like salient regions when up-sampled to the original image size that might not be descriptive enough. In Fig. \ref{fig:saliency} input saliency maps are depicted for both baseline and SSBVER models with ResNet50\_IBN architecture and for a pair of images selected from VeRi dataset. It is observed that the baseline model mainly focused on the front bumper of the vehicle and slightly attended the side as it a shared portion between the two images. However, SSBVER asserts more attention on discriminative cues such as the white napkin box on the dashboard and the car's side skirt. Also the similarity scores obtained for this image pair via baseline and SSBVER models are $0.96$ and $0.98$ respectively. This examples qualitatively explains the reduction in the $L_2$ distance of features extracted from positive pairs and the intra-class compactness discussed in section \ref{subsec:intra-class-compcatness} and Fig. \ref{fig:feat_distant_dist}. While the baseline is a strong re-id model, incorporation of self-supervision via knowledge distillation and self-training helps to learn more discriminative and locally-distinguishable information without employing any explicit and computation-demanding attention mechanism.

\subsection{Comparison with the state-of-the-art}
\label{sec:sota}
In this section, we compare SSBVER with ResNet50\_IBN backbone architecture against recent works on vehicle re-id in terms of both evaluation and efficiency metrics. The reason we chose ResNet50\_IBN for SSBVER compared to ResNet50, ViT, SWIN, and ConvNext backbones is that it maintains a comparatively high level of accuracy on all benchmarks while keeping the inference speed and resource utilization low as reported in Tables \ref{tab:backbone_stats}, \ref{tab:veri_results}, \ref{tab:vehicleid_results}, and \ref{tab:veriwild_results}. Table \ref{tab:sota_comp} reports the result of comparison with recent methods. In a first glance, HRCN \cite{zhao2021heterogeneous} appears to be the superior model in terms of evaluation metrics; however, it takes $10.84$ milliseconds to compute $3584$-dimensional embeddings. Using this model in a multi-camera tracking system that is intended to track hundreds or thousands of vehicles in real-time would be overwhelming and can exhaust all the resources quickly. SSBVER on the other hand, is a very simple and light weight approach that does not rely on additional annotations, and has a performance that is comparable to HRCN and higher than other computationally expensive alternatives such as TransReID \cite{he2021transreid} and PVEN \cite{meng2020parsing} in particular. We emphasize that SSBVER relies on only a single backbone feature extractor model that can be easily modified to meet resource constraints depending on the application. Table \ref{tab:sota_comp} also highlights that efficiency metrics are mainly ignored in the community as we had to measure them by re-implementing the corresponding works. Lastly, we  point out that reported numbers for Speed (ms/image) should be considered for relative comparison. These can be further reduced depending on the hardware and adopting inference time optimizations libraries such as NVIDIA TensorRT\footnote{\url{https://developer.nvidia.com/tensorrt}}. 

\begin{table}[t!]
    \caption{Comparison with recent state-of-the-arts methods. Note that $^*$ denotes the number is not reported in the original paper and is computed by implementing the corresponding work or adopting the official repository upon availability.}
    \centering
    \label{tab:sota_comp}
    \resizebox{1.\columnwidth}{!}{
    \begin{tabular}{c|c|c|c|c|c|c|c|c|c|c|c|c|c}
    \hline
        \multicolumn{1}{c}{\multirow{4}{*}{Method}} & \multicolumn{9}{||c|}{Evaluation Metrics} & 
        \multicolumn{4}{||c}{Efficiency Metrics}
        \\
         \cline{2-14}
         \multicolumn{1}{c}{} & \multicolumn{3}{||c|}{VeRi} &
         \multicolumn{3}{|c|}{VehicleID (L)} &
         \multicolumn{3}{|c|}{VeRiWild (S)} &
         \multicolumn{1}{||c|}{\multirow{3}{*}{Params}} &
         \multicolumn{1}{|c|}{\multirow{3}{*}{Dims}} &
         \multicolumn{1}{|c|}{\multirow{3}{*}{Speed}} &
         \multicolumn{1}{|c}{\multirow{3}{*}{Memory}} \\
         \cline{2-10}
         \multicolumn{1}{c}{} & \multicolumn{1}{||c|}{\multirow{2}{*}{mAP}} & \multicolumn{2}{|c|}{CMC} &
         \multicolumn{1}{|c|}{\multirow{2}{*}{mAP}} & \multicolumn{2}{|c|}{CMC} &
         \multicolumn{1}{|c|}{\multirow{2}{*}{mAP}} & \multicolumn{2}{|c|}{CMC} & \multicolumn{1}{||c|}{} & \multicolumn{1}{|c|}{} & \multicolumn{1}{|c|}{} & \multicolumn{1}{|c}{} \\
         \cline{3-4} \cline{6-7} \cline{9-10}
         \multicolumn{1}{c}{} & \multicolumn{1}{||c}{} & \multicolumn{1}{|c|}{@1} & \multicolumn{1}{|c|}{@5} & \multicolumn{1}{|c}{} & \multicolumn{1}{|c|}{@1} & \multicolumn{1}{|c|}{@5} & \multicolumn{1}{|c}{} & \multicolumn{1}{|c|}{@1} & \multicolumn{1}{|c|}{@5} & \multicolumn{1}{||c|}{(M)} & \multicolumn{1}{|c|}{} & \multicolumn{1}{|c|}{(ms/image)} & \multicolumn{1}{|c}{(MB)} \\
         \hline \hline
         TransReID\cite{he2021transreid} & 
         \multicolumn{1}{||c|}{81.4} & 
         \multicolumn{1}{|c|}{96.8} & 
         \multicolumn{1}{|c|}{98.4} & 
         \multicolumn{1}{|c|}{84.9} & 
         \multicolumn{1}{|c|}{78.7} & 
         \multicolumn{1}{|c|}{93.2} & 
         \multicolumn{1}{|c|}{$81.2^*$} & 
         \multicolumn{1}{|c|}{$92.3^*$} & 
         \multicolumn{1}{|c|}{$98.0^*$} &
         \multicolumn{1}{||c|}{$101^*$} &
         \multicolumn{1}{|c|}{$3840^*$} &
         \multicolumn{1}{|c|}{$5.51^*$} &
         \multicolumn{1}{|c}{$423^*$}\\
         GFDIA \cite{li2021self} & \multicolumn{1}{||c|}{81.0} & \multicolumn{1}{|c|}{96.7} & \multicolumn{1}{|c|}{98.6} & \multicolumn{1}{|c|}{-} & 
         \multicolumn{1}{|c|}{80.0} &
         \multicolumn{1}{|c|}{93.7} &
         \multicolumn{1}{|c|}{-} & 
         \multicolumn{1}{|c|}{-} & 
         \multicolumn{1}{|c|}{-} &
         \multicolumn{1}{||c|}{$34.7^*$} &
         \multicolumn{1}{|c|}{$4096^*$} &
         \multicolumn{1}{|c|}{$5.74^*$} &
         \multicolumn{1}{|c}{$173^*$}\\
         SAVER \cite{khorramshahi2020devil} & \multicolumn{1}{||c|}{79.6} & \multicolumn{1}{|c|}{96.4} & \multicolumn{1}{|c|}{98.6} & \multicolumn{1}{|c|}{82.9} & 
         \multicolumn{1}{|c|}{75.3} &
         \multicolumn{1}{|c|}{88.3} &
         \multicolumn{1}{|c|}{80.9} & 
         \multicolumn{1}{|c|}{94.5} & 
         \multicolumn{1}{|c|}{98.1} &
         \multicolumn{1}{||c|}{$31^*$} &
         \multicolumn{1}{|c|}{$2048^*$} &
         \multicolumn{1}{|c|}{$5.06^*$} &
         \multicolumn{1}{|c}{$178^*$}\\
         
         EVER \cite{Peri_2020_CVPR_Workshops} &
         \multicolumn{1}{||c|}{80.4} &
         \multicolumn{1}{|c|}{95.8} &
         \multicolumn{1}{|c|}{97.9} &
         \multicolumn{1}{|c|}{84.3} & 
         \multicolumn{1}{|c|}{78.4} &
         \multicolumn{1}{|c|}{92.3} &
         \multicolumn{1}{|c|}{80.7} & 
         \multicolumn{1}{|c|}{93.7} & 
         \multicolumn{1}{|c|}{97.8} &
         \multicolumn{1}{||c|}{$23.5^*$} &
         \multicolumn{1}{|c|}{$2048^*$} &
         \multicolumn{1}{|c|}{$4.55^*$} &
         \multicolumn{1}{|c}{$122^*$}\\
         
         %DFLNET \cite{bai2020disentangled} & 
         %\multicolumn{1}{||c|}{80.9} & 
         %\multicolumn{1}{|c|}{97.1} & 
         %\multicolumn{1}{|c|}{\textbf{99.0}} & 
         %\multicolumn{1}{|c|}{83.2} & 
         %\multicolumn{1}{|c|}{79.1} &
         %\multicolumn{1}{|c|}{92.8} &
         %\multicolumn{1}{|c|}{83.1} & 
         %\multicolumn{1}{|c|}{94.8} & 
         %\multicolumn{1}{|c|}{98.0} &
         %\multicolumn{1}{||c|}{-} &
         %\multicolumn{1}{|c|}{-} &
         %\multicolumn{1}{|c|}{-} &
         %\multicolumn{1}{|c}{-}\\
         
         PVEN \cite{meng2020parsing} & 
         \multicolumn{1}{||c|}{79.5} & 
         \multicolumn{1}{|c|}{95.6} & 
         \multicolumn{1}{|c|}{98.4} & 
         \multicolumn{1}{|c|}{-} & 
         \multicolumn{1}{|c|}{77.8} &
         \multicolumn{1}{|c|}{92.0} &
         \multicolumn{1}{|c|}{79.8} & 
         \multicolumn{1}{|c|}{94.0} & 
         \multicolumn{1}{|c|}{98.0} &
         \multicolumn{1}{||c|}{$59.2^*$} &
         \multicolumn{1}{|c|}{$10240^*$} &
         \multicolumn{1}{|c|}{$11.79^*$} &
         \multicolumn{1}{|c}{$603^*$}\\
         
         HRCN \cite{zhao2021heterogeneous} & 
         \multicolumn{1}{||c|}{\textbf{83.1}} & 
         \multicolumn{1}{|c|}{\textbf{97.3}} & 
         \multicolumn{1}{|c|}{\textbf{98.9}} & 
         \multicolumn{1}{|c|}{$\textbf{85.9}^*$} & 
         \multicolumn{1}{|c|}{$\textbf{79.5}^*$} &
         \multicolumn{1}{|c|}{$\textbf{94.8}^*$} &
         \multicolumn{1}{|c|}{\textbf{85.2}} & 
         \multicolumn{1}{|c|}{93.8} & 
         \multicolumn{1}{|c|}{$98.3^*$} &
         \multicolumn{1}{||c|}{$55.4^*$} &
         \multicolumn{1}{|c|}{$3584^*$} &
         \multicolumn{1}{|c|}{$10.84^*$} &
         \multicolumn{1}{|c}{$260^*$}\\
         
         \hline \hline SSBVER & 
         \multicolumn{1}{||c|}{82.1} & 
         \multicolumn{1}{|c|}{97.1} & 
         \multicolumn{1}{|c|}{98.4} & 
         \multicolumn{1}{|c|}{84.8} & 
         \multicolumn{1}{|c|}{78.9} &
         \multicolumn{1}{|c|}{92.6} &
         \multicolumn{1}{|c|}{82.6} & 
         \multicolumn{1}{|c|}{\textbf{95.1}} & 
         \multicolumn{1}{|c|}{\textbf{98.5}} &
         \multicolumn{1}{||c|}{\textbf{23.5}} &
         \multicolumn{1}{|c|}{\textbf{2048}} &
         \multicolumn{1}{|c|}{\textbf{4.55}} &
         \multicolumn{1}{|c}{\textbf{122}}\\
    \end{tabular}}
\end{table}

\section{Ablation Studies}
\label{sec:ablation}
Here we compare the cross entropy loss employed in the self-supervision objective in Eq. \ref{eq:dino}, used to encourage the student to match the teacher model output, against an alternative. In addition, we study how the performance of SSBVER varies upon changing the number of local crops in the set $V_l$.

\begin{figure}[]
    \centering
    \includegraphics[width=0.44\textwidth]{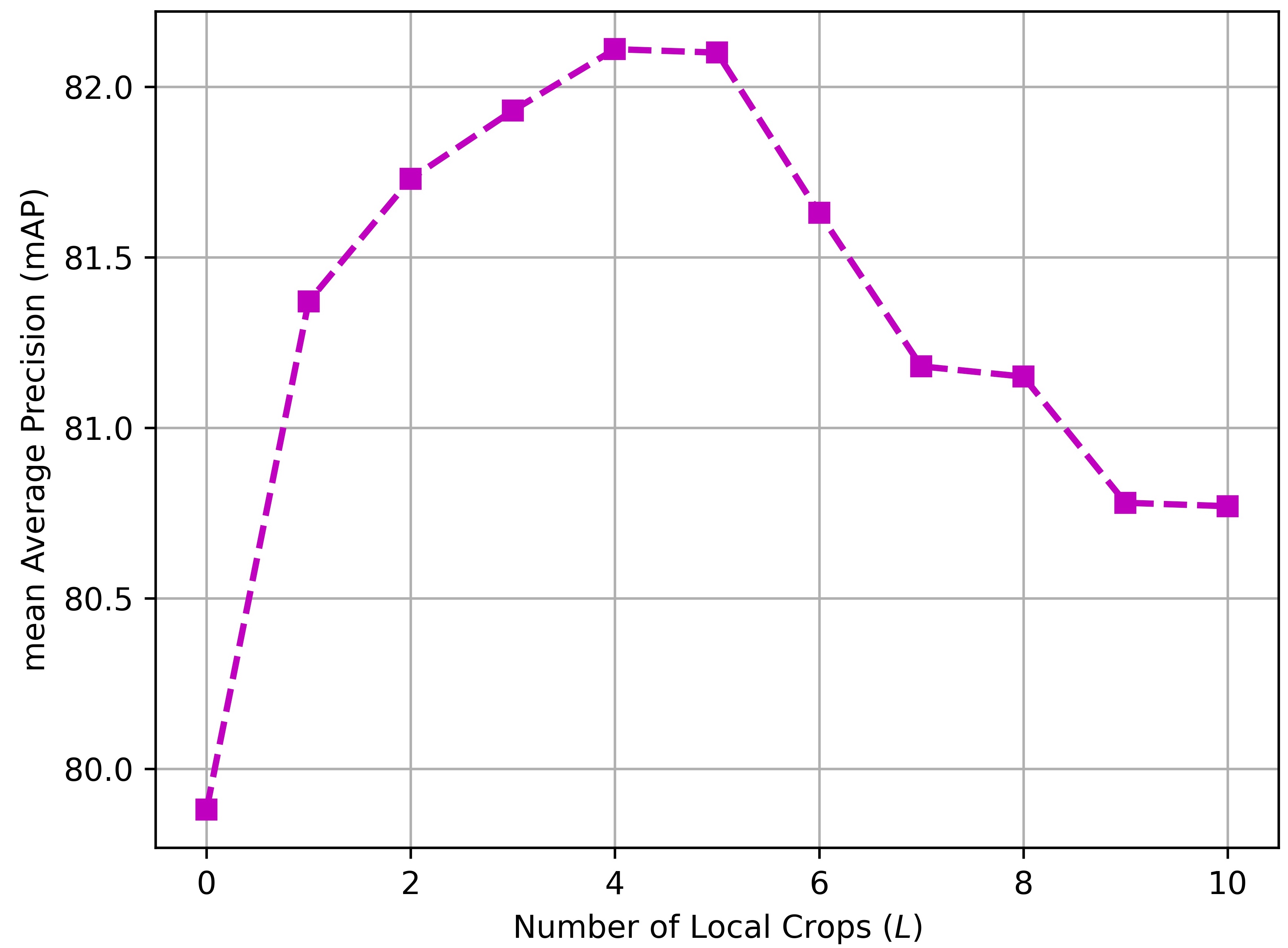}
    \caption{Impact of the number of local crops $L$ on the mean Average Precision (mAP). Here we use ResNet50\_IBN backbone architecture and VeRi dataset.}
    \label{fig:L_ablation}
\end{figure}

First we replace the cross entropy objective function in Eq. \ref{eq:dino} between the MLP outputs of student and teacher branches with the $L_2$ norm of their difference which is often referred to as Root Mean Squared Error (RMSE) loss. Therefore, the self-supervised loss $\mathcal{L}_s$ is calculated by:
\begin{equation}
    \label{eq:dino_MSE}
    \centering
    \mathcal{L}_s = \sum_{\scalebox{0.6}{$I\in V_g(I)$}} \sum_{\begin{array}{c}
                            \scalebox{0.6}{$I^{'}\in V_g(I) \cup V_l(I)$} \\
                            \scalebox{0.6}{$I^{'} \neq I$} \end{array}} {||g_s(f_s(I^{'})) - g_t(f_t(I))||}_2
\end{equation}
Table \ref{tab:objective_comparison} presents the result of this comparison for VeRi dataset and ResNet50\_IBN architecture. It is seen that minimizing the cross entropy between the predictions of the student model and the teacher model results in a higher performance compared to directly minimizing their RMSE. This observation is consistent with the findings of authors in the original DINO work \cite{caron2021emerging} where they attempt to learn purely self-supervised features from the scratch while we learn vehicle representations in a hybrid manner, \emph{i.e.} supervised learning (Re-ID Head) along with self-supervised learning (SSL Head).
\begin{table}[t!]
    \centering
    \caption{Comparison between different choices of objective functions for self-supervision enforcement. Numbers are reported for VeRi dataset and ResNet50\_IBN architecture.}
    \label{tab:objective_comparison}
    \resizebox{0.6\columnwidth}{!}{
    \begin{tabular}{c|c|c|c}
    \hline
        \multicolumn{1}{c|}{\multirow{2}{*}{Objective Function}} & \multicolumn{3}{||c}{Evaluation Metrics}  \\
        \cline{2-4}
        \multicolumn{1}{c|}{} & \multicolumn{1}{||c}{mAP(\%)} & \multicolumn{1}{|c|}{CMC@1(\%)} & \multicolumn{1}{|c}{CMC@5(\%)} \\
        \hline \hline 
        \multicolumn{1}{c|}{RMSE} & \multicolumn{1}{||c|}{$80.80$} &\multicolumn{1}{|c|}{$96.54$} & \multicolumn{1}{|c}{$98.39$}\\
        \hline
        \multicolumn{1}{c|}{Cross Entropy} & \multicolumn{1}{||c|}{$\textbf{82.11}$} & \multicolumn{1}{|c|}{$\textbf{97.08}$} & \multicolumn{1}{|c}{$\textbf{98.45}$}\\
        \hline
    \end{tabular}
    }
\end{table}
Additionally, we would like to understand how the performance of the SSBVER varies with respect to the number of local crops $L$ that student model observes for each sample during training. Fig. \ref{fig:L_ablation} demonstrates the re-id performance in terms of mAP as a function of $L$ for the case of VeRi dataset and ResNet50\_IBN backbone architecture. In Fig. \ref{fig:L_ablation} there is a significant jump from the baseline ($L=0$) to when apply self-supervision with only $L=1$ local crop. The maximum mAP occurs for $L=4$ crops and as it is increased, the performance starts to drop. The reduction in performance for large number of local crops can be attributed to observing more regions of a target image by student network and spending less effort to match the output of teacher. This in turn leads to learning less discriminative representations. Nevertheless, we should note that performance is considerably higher than the baseline model.

\section{Conclusions}
\label{sec:conclusion}
In this work we present the SSBVER vehicle re-identification model that is boosted by self-supervision through self-training and knowledge distillation during training time. SSBVER imitates the computationally expensive attention mechanisms which are proved to be critical for obtaining robust vehicle representations. SSBVER yields performance improvements consistently on publicly available benchmarks irrespective to the choice of the backbone DNN architecture. In contrast to vehicle re-id alternatives, our proposed approach only requires a forward pass of a single DNN model architecture without any additional overhead. As vehicle re-identification technology becomes more mature, its large-scale deployment in applications such as City-Scale Multi-Camera Tracking seems to be reachable more than ever. As a result, additional emphasis should be directed towards efficiency metrics such as throughput, dimensionality of output embeddings, number of parameters and hence the GPU memory footprint which are often overlooked in the community. The importance of these metrics becomes quite evident in real-time and at-scale applications where the amount of data to be processed and managed is overwhelming. SSBVER obtains performance on par to state-of-the-art approaches in spite of being computationally far less demanding. Therefore, we advocate for efficiency metrics and hope this work motivates further research to develop efficient and lightweight models suited for at-scale applications.

%\bibliographystyle{splncs04}
%\bibliography{egbib}

\end{document}